\documentclass{bmvc2k}

\title{\titletext}

\addauthor{Bj\"orn Michele}{bjoern.michele@univ-ubs.com}{1,2}
\addauthor{Alexandre Boulch}{alexandre.boulch@valeo.com}{1}
\addauthor{Gilles Puy}{gilles.puy@valeo.com}{1}
\addauthor{Tuan-Hung Vu}{tuan-hung.vu@valeo.com}{1}
\addauthor{Renaud~Marlet}{renaud.marlet@valeo.com}{1,3}
\addauthor{Nicolas Courty}{nicolas.courty@univ-ubs.com}{2}

\addinstitution{
 valeo.ai,\\ Paris, France
}
\addinstitution{
CNRS, IRISA, Univ.\ Bretagne Sud, Vannes, France\\
}
\addinstitution{
LIGM, Ecole des Ponts, Univ Gustave Eiffel, CNRS,\\ Marne-la-Vall\'ee, France
}

\runninghead{Michele et al.}{Improving Multimodal Distillation for 3D Sem. Seg. UDA}

\usepackage{graphicx}
\usepackage{amsmath}
\usepackage{amssymb}

\usepackage[utf8]{inputenc} 
\usepackage[T1]{fontenc}    
\usepackage{url}            
\usepackage{booktabs}       
\usepackage{amsfonts}       
\usepackage{nicefrac}       
\usepackage{microtype}      
\usepackage[dvipsnames]{xcolor}         

\usepackage{amssymb}
\usepackage{pifont}
\usepackage{colortbl}
\usepackage{rotating}
\usepackage{scalerel}
\usepackage[normalem]{ulem}

\usepackage[capitalize]{cleveref}
\crefname{section}{Sec.}{Secs.}
\Crefname{section}{Section}{Sections}
\Crefname{table}{Table}{Tables}
\crefname{table}{Tab.}{Tabs.}
\Crefname{figure}{Figure}{Figures}
\crefname{figure}{Fig.}{Figs.}

\usepackage{multirow}

\usepackage{bm}
\usepackage{adjustbox}
\usepackage{array}
\usepackage{enumitem}

\newcolumntype{R}[2]{%
    >{\adjustbox{angle=#1,lap=1.3\width-(#2)}\bgroup}%
    l%
    <{\egroup}%
}

\usepackage[linesnumbered,ruled,vlined]{algorithm2e}
\usepackage{algpseudocode}
\makeatletter 
\@namedef{ver@everyshi.sty}{}
\makeatother
\usepackage{tikz}
\usetikzlibrary{positioning,shapes}

\definecolor{MyGreen}{RGB}{0, 180, 0}
\definecolor{MyRed}{RGB}{180, 0, 0}
\definecolor{MyBlue}{RGB}{30, 0, 180}
\definecolor{MyGrey}{RGB}{82.75, 82.75, 82.75}

\definecolor{skcar}{RGB}{100,150,245}
\definecolor{skbicycle}{RGB}{100,230,245}
\definecolor{skmotorcycle}{RGB}{30,60,	150}
\definecolor{sktruck}{RGB}{80,30,180}
\definecolor{skotherv}{RGB}{0, 0, 255}
\definecolor{skpedestrian}{RGB}{255,30,30}
\definecolor{skdrivable}{RGB}{255, 0,255}
\definecolor{sksidewalk}{RGB}{75, 0,75}
\definecolor{skterrain}{RGB}{150, 240, 80}
\definecolor{skvegetation}{RGB}{0, 175, 0}
\definecolor{skbuilding}{RGB}{255, 200, 0}

\definecolor{tsne_source}{rgb}{0.86, 0.3712, 0.33999999999999997}
\definecolor{tsne_target}{rgb}{0.33999999999999997, 0.8287999999999999, 0.86}

\makeatletter
\renewcommand\paragraph{\@startsection{paragraph}{4}{\z@}%
    {0.7ex \@plus0.5ex \@minus.2ex}%
    {-1em}%
    {\normalfont\normalsize\bfseries}}
\makeatother

\newcommand{\cmark}{{\textcolor{MyGreen}{\ding{51}}}}%
\newcommand{\xmark}{{\textcolor{MyRed}{\ding{55}}}}%

\def \method{MuDDoS\xspace}

\def \lidar{lidar\xspace}

 \newcommand{\titletext}{
 Improving Multimodal Distillation for 3D Semantic Segmentation under Domain Shift 
 }

\newcommand{\back}[1]{\it\textcolor{black!30}{#1}}

\usepackage{stmaryrd}
\usepackage{trimclip}

\makeatletter
\DeclareRobustCommand{\shortto}{%
  \mathrel{\mathpalette\short@to\relax}%
}

\newcommand{\short@to}[2]{%
  \mkern2mu
  \clipbox{{.5\width} 0 0 0}{$\m@th#1\vphantom{+}{\shortrightarrow}$}%
  }
\makeatother

\newcommand{\ra}[1]{\renewcommand{\arraystretch}{#1}}

\begin{document}

\maketitle


\begin{abstract}
Semantic segmentation networks trained under full supervision for one type of lidar fail to generalize to unseen lidars without intervention. 
To reduce the performance gap under domain shifts, a recent trend is to leverage vision foundation models (VFMs) providing robust features across domains.
In this work, we conduct an exhaustive study to identify recipes for exploiting VFMs in unsupervised domain adaptation for semantic segmentation of lidar point clouds. 
Building upon unsupervised image-to-lidar knowledge distillation, our study reveals that: (1) the architecture of the lidar backbone is key to maximize the generalization performance on a target domain; (2) it is possible to pretrain a single backbone once and for all, and use it to address many domain shifts; (3) best results are obtained by keeping the pretrained backbone frozen and training an MLP head for semantic segmentation.
The resulting pipeline achieves state-of-the-art results in four widely-recognized and challenging settings. The code will be available~at:~\begingroup \hypersetup{urlcolor=red}{\href{https://github.com/valeoai/muddos}{github.com/valeoai/muddos}}.\endgroup
\end{abstract}
    
\section{Introduction}
\label{sec:intro}

Understanding scenes at 3D level is key for applications like autonomous driving or robotics. In particular, the semantic segmentation of lidar scans is valuable high-level information that autonomous vehicles can rely upon, e.g., {for trajectory planning}. However, state-of-the-art networks for semantic segmentation require a large amount of costly annotated training data to {achieve }good performance, limiting their deployment in new {environments} or when changing sensors.
Domain adaptation (DA) addresses this problem by adapting a network trained on a labeled source domain to a new target domain. Unsupervised domain adaptation (UDA), in particular, conducts this adaptation without using any label on the target domain.

Vision foundation models (VFMs), trained on web-scale datasets, provide features that can be used off-the-shelf for a wide variety of tasks, and are robust to strong domain shifts \cite{oquab2023dinov2, wei2024stronger}. 
In addition, recent works like \cite{sautier2022image,mahmoud2023self,liu2023segment,puy2024three,superflows} show that knowledge distillation of VFMs into 3D backbones can provide robust 3D features. In this work, we build upon these distillation methods and conduct an exhaustive study to identify key recipes for exploiting VFMs for 3D UDA on autonomous driving datasets in the best possible way. Notably, our study identifies practices that allow us to outperform
by more than 15 mIoU points the previous state of the art in multimodal unsupervised domain adaptation for 3D lidar semantic segmentation.

We structure our study along four axes. First, we analyze the effect of different
architectural choices in the 3D backbone to improve its robustness to domain gaps. Second, we benchmark different VFMs to identify the most appropriate for our task. Third, we evaluate different downstream training recipes on source datasets to get the best models on target datasets. 
Finally, we study how the pretraining datasets used for distillation influence performance under domain shifts.

\begin{figure*}[t]
    \includegraphics[width=\linewidth]{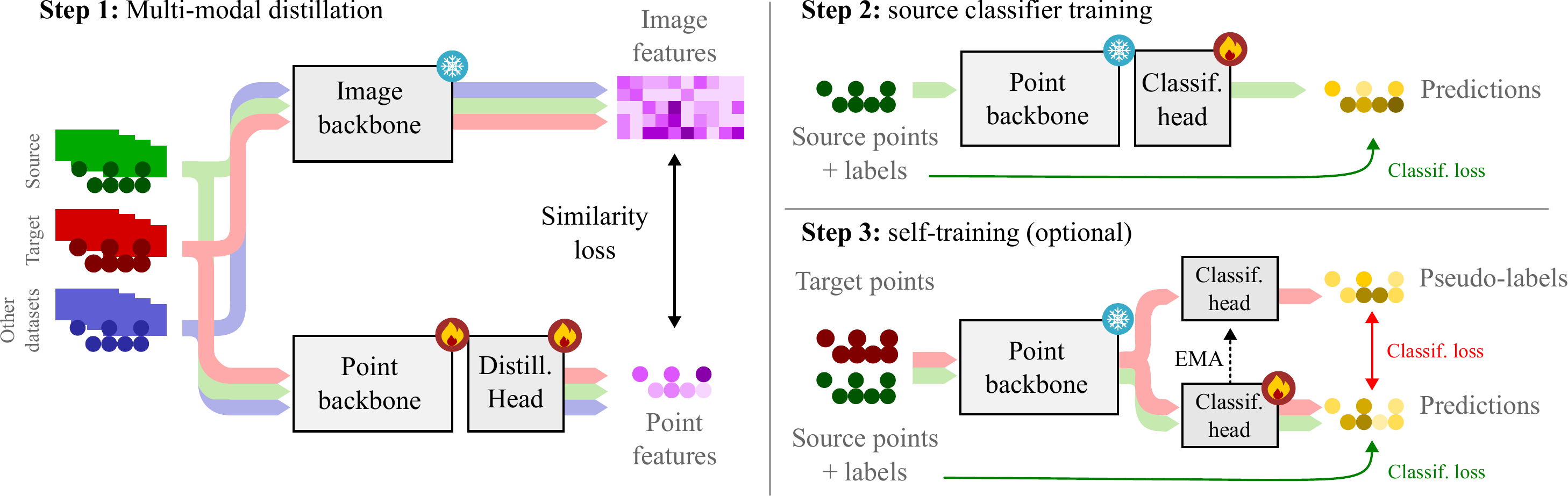}
    \caption{
        \textbf{Overview of the multimodal distillation pipeline for 3D domain adaptation.} With \method, adapting from an annotated source dataset to an unannotated target dataset, operating in three steps. \textbf{Step 1} is a 2D-to-3D distillation using a frozen visual foundation model (DINOv2) to obtain aligned 3D representations on all datasets.
        \textbf{Step~2} trains a classification head with source labels. The backbone is frozen to prevent the 3D representations from drifting away and to maintain a good performance on the target dataset.
        \textbf{Step~3} is a prediction refinement using self-training obtained via a classical teacher-student scheme.
        }

    \label{fig:pipeline}
\end{figure*}

Our insights are listed below.
\setlist{nolistsep}
\begin{itemize}[noitemsep]
    \item 
    \textit{Backbone}. While we are able to surpass competing methods with MinkowskiUNet \cite{choy20194d} (MUNet), which is the default choice of backbone in the DA literature, we advocate the use of more recent networks, which compete with the best adapted MUNet without even leveraging any domain adaptation technique. Further improvements can be achieved by scaling the capacity of the backbone. We also notice that the choice of normalization layers has a high impact on the generalization capabilities of the lidar backbone, where layernorms perform better than batchnorms. Besides, we observe that the use of lidar intensity as input feature is most often detrimental {for adaptation}.
    \item 
    \textit{Pretraining by distillation}. We note that ViTs pretrained with DINOv2 provide more robust features than those provided by SAM \cite{zou2023segment}. Moreover, we show that distillation can be performed \textit{once for all} on a combination of multiple datasets. This contrasts with existing techniques where a new full training is done for each new pair of source and target datasets.
    \item 
    \textit{Downstream training}. Generalization capabilities are preserved when the backbone is kept frozen after distillation and a classification head is trained for the downstream task of semantic segmentation. Better results are obtained when the distillation and semantic segmentation training are done consecutively, rather than jointly. 
\end{itemize}
The best technique resulting from this study obtains SOTA results with large margins. We call it \method, for \textbf{Mu}ltimodal \textbf{D}istillation for 3D Semantic Segmentation under \textbf{Do}main \textbf{S}hifts. An overview of the pipeline can be seen in~\cref{fig:pipeline}.

\section{Related work}

\paragraph{Monomodal UDA for semantic segmentation.}
Taking inspiration from the problem of UDA for images, several works adapted these techniques to lidar scenes such as adversarial training via projection of the points to image-like representations~\cite{zhao2021epointda, barrera2021cycle, debortoli2021adversarial, jiang2021lidarnet, li2023adversarially}, mixing source and target data~\cite{saltori2022cosmix, kong2023conda}, or using both adversarial training and mixing strategies~\cite{yuan2024density}.
Geometric methods target the specificity of the domain gap induced by different sensors, e.g., variable acquisition patterns or location on the vehicle.
For example, the source point clouds can be up/down-sampled to resemble the target point clouds~\cite{wei2022lidar}, possibly with self-ensembling~\cite{shaban2023lidar}.
Along the same line, a reconstruction of the underlying sensor-agnostic surface can be used as a unified representation space~\cite{yi2021complete} or as a regularization loss~\cite{michele2024saluda}. In addition to the vanilla UDA settings, several others have studied source-free UDA~\cite{michele2024ttyd}, online-adaptation~\cite{saltori2022gipso} and generalization~\cite{kim2023single, Sanchez_2023_ICCV, saltori2023walking, zhao2024unimix, jiang2024dg, park2024rethinking, sanchez2025cola}.

\paragraph{Multimodal UDA for semantic segmentation.}

Automotive datasets are usually acquired with several synchronized sensors, including lidars and cameras~\cite{lidarseg_nuscenes, geiger2012cvpr, behley2019iccv,Ettinger_2021_ICCV,xiao2021pandaset}.
Using the sensor calibration and their relative position, points can be projected in the accompanying images. Thus, it makes it possible to do a multimodal domain adaptation that leverages the image as well as the lidar modalities.  

\textit{Training the image features extractor.}
The pioneering work of xMUDA~\cite{jaritz2020xmuda, jaritz2022cross} introduced the task of multimodal domain adaptation for semantic segmentation using both \lidar and image data. Their approach involved training separate classifiers for each modality — one for images and one for point clouds — while enforcing consistency through a KL-divergence loss between the semantic predictions of corresponding \lidar points and image pixels. 
At inference, the best performance was achieved by fusing the predictions of both classifiers, leveraging the complementarity of 2D and 3D data. Though effective, this approach requires precisely calibrated \lidar-camera data at test time and limits predictions to regions visible in both modalities. 
For example, in SemanticKITTI~\cite{geiger2012cvpr,behley2019iccv} only a front-view image is available, in Waymo~\cite{Ettinger_2021_ICCV} the rear part of the lidar scan is not covered by the images, and in nuScenes~\cite{caesar2020nuscenes}, even though using a complete ring of cameras, only 48\% of the points have a reprojection in 2D~\cite{ovsep2024better}.
Several works have built upon xMUDA to enhance multi-modal domain adaptation~\cite{LIU2021211, peng2021sparse, cardace2023exploiting, zhang2023mx2m}. Furthermore, multimodal information has been leveraged for test-time adaptation~\cite{shin2022mm, weijler2024ttt}.

\textit{Using a vision foundation model.} VFMs are trained with distinct objectives, resulting in unique characteristics. SAM~\cite{zou2023segment} generates high-quality masks of objects instances in image space. DINOv2~\cite{oquab2023dinov2} focuses on extracting high-quality semantically-coherent features. In multimodal domain adaptation, methods leverage these models' specific strengths.
Several works~\cite{peng2025learning, xu2024visual, cao2024mopa} rely on the ability of SAM to extract object instances. Adapt-SAM~\cite{peng2025learning} uses it to produce instance masks in each domain, selecting some instances in each domain, and adding these instances in the point cloud of the other domain. An image-to-lidar distillation loss using SAM features is also used while training the lidar backbone for semantic segmentation. This approach is the closest to ours as it leverages image-to-lidar distillation during training and performs inference using only the \lidar data and predicts semantic labels for the entire $360^{\circ}$ point cloud (and not only for points with an image projection).

\paragraph{Image-to-lidar distillation.}

Our method leverages image-to-lidar distillation techniques. Among such techniques, a first set of techniques distill the knowledge of vision language models such as CLIP \cite{radford2021learning} to make open-vocabulary tasks on lidar data possible~\cite{chen2023clip2scene, peng2023openscene, 
 zhang2022pointclip,zhu2023pointclip}. 
Another set of techniques uses VFM as teacher such as \cite{caron2021emerging,oquab2023dinov2} to pretrain lidar backbones without supervision \cite{sautier2022image, liu2021learning,  mahmoud2023self, mahmoud24distill, superflows}. This pretraining stage permits to reach better performance on downstream tasks such as semantic segmentation or object detection. In this paper, we build upon ScaLR \cite{puy2024three} which detailed several techniques to improve knowledge distillation.
%
\section{Study: How to get the best out of VFMs for 3D UDA?}

\paragraph{Experimental setup.}

We conduct our study using three datasets:
nuScenes~\cite{lidarseg_nuscenes}~(N), SemanticKITTI~\cite{geiger2012cvpr, behley2019iccv}~(K) and Waymo Open Dataset~\cite{Ettinger_2021_ICCV}~(W).
The pairs of source and target datasets considered, which are common in the literature, are:  N$\shortto$K, K$\shortto$N, N$\shortto$W and W$\shortto$N. A different lidar sensor is used in each dataset. In nuScenes, the lidar used has $32$ beams, vs. $64$ beams in SemanticKITTI and Waymo. All 3 sensors also have a different horizontal resolution. These variations make the settings especially difficult. The precise class mappings between the source and target datasets are provided in the supplementary material, as well as the training protocols (augmentation, losses and optim. parameters). See supp.\,mat.\ B and A.

\subsection{Choice of lidar backbone architecture}
\label{sec:arch}

A common choice of backbone in the lidar UDA literature is MinkowskiUNet \cite{choy20194d} (MUNet). We show that this default choice actually limits our ability to reach high performance. We advocate for the use of other architectures that are more robust to domain shifts.

We study different choices of lidar backbone architecture. 
The architectures considered are MinkowskiUNet~\cite{choy20194d} (MUNet) and WaffleIron~\cite{puy23waffleiron} (WI, with 256 and 768 channel variants).
We also use the MUNet from ~\cite{saltori2022cosmix, peng2025learning} in \cref{tab:experiments_main_comparison}. 
In this first part of the study, all backbones are trained from scratch for semantic segmentation on source datasets \textit{without any pretraining phase}. 
After training, the backbone is directly tested on target datasets.
We monitor the effect of: 
(a)~using the intensity as input feature; 
(b)~changing batchnorms to layernorms; 
(c)~increasing the capacity of the backbones. 
The results are presented in \cref{tab:experiments_uni_modal_ingredients}.

\begin{table}
\parbox{.485\linewidth}{
\scriptsize
\begin{center}
\setlength{\tabcolsep}{3.5pt}
\begin{tabular}{l@{\hskip 1pt}cc|cccc@{}}
\toprule
Backbone
    & Int.
    & Norm 
    & \scriptsize N$\shortto$K
    & \scriptsize K$\shortto$N
    & \scriptsize N$\shortto$W
    & \scriptsize W$\shortto$N 
\\
\midrule
\multirow{4}{*}{MUNet}
    & \cmark
    & BN   
    & 30.1
    & 22.0 
    & 28.3  
    & 30.9
\\
    & \xmark
    & BN   
    & 32.3 
    & 49.9
    & 31.9 
    & 46.2
\\
    & \cmark
    & LN
    & 35.9 
    & 21.6 
    & 27.9 
    & 31.5 
\\
    & \xmark
    & LN
    & 38.1
    & 49.5 
    & 33.1
    & 44.7 

\\
\midrule
\multirow{4}{*}{WI-256}
    & \cmark      
    & BN 
    & 16.0 
    & 48.0
    & 20.2 
    & 59.7 
\\
    & \xmark      
    & BN
    & 33.8 
    & 52.9 
    & 19.9  
    & 60.1 
\\
    & \cmark 
    & LN
    & 22.2 
    & 50.6 
    & 40.9 
    & 63.2 
\\
    & \xmark 
    & LN
    & 41.1 
    & 49.8 
    & 35.3 
    & 61.8 
\\
\midrule
WI-768 
    & \xmark 
    & LN 
    & 44.6 
    & \bf 55.1 
    & 37.1  
    & \bf 64.6 
\\
\midrule
\rowcolor{gray!20}
\multicolumn{7}{l}{\it State of the art in lidar multi-modal UDA.}
\\
\shortstack[c]{MUNet w.\,\cite{peng2025learning}}
    & \cmark 
    & BN  
    & \bf48.5 
    & 42.9 
    & \bf 44.9 
    & 48.2\\
\bottomrule
\end{tabular}
\end{center}
\vspace{-0.2cm}
\caption{\textbf{Generalization capability of different backbones} and architecture choices, intensity (Int.) or normalization layer (BN, LN). Trained on source, evaluated on target.}
\label{tab:experiments_uni_modal_ingredients}
}
\hfill
\parbox{.485\linewidth}{
\scriptsize
\begin{center}
\setlength{\tabcolsep}{10pt}
\vspace{-0.17cm}
\begin{tabular}{ll|cc}
\toprule
Backbone& VFM & N$\shortto$K & K$\shortto$N\\
\midrule
\multirow{2}{*}{WI-256}
    & SAM ViT-L & 42.5  & 48.6  \\
    & DiNOv2 ViT-L & \bf 46.6 & \bf 59.7     \\
\bottomrule
\end{tabular}
\end{center}
\caption{
\textbf{Effect of the VFM} used for feature distillation with ScaLR.
}
\label{tab:experiments_sam_vs_dino}

\begin{center}
\ra{1.1}
\setlength{\tabcolsep}{4pt}
\begin{tabular}{lc|cc}
\toprule
\multirow{1}{*}{\shortstack[l]{Downstream training}}
    & \multirow{1}{*}{Intensity}
    & N$\shortto$K
    & K$\shortto$N
\\
\midrule
Frozen Back. + Linear {(a)}
    & \xmark
    & 43.4
    & \underline{60.1}
\\
Frozen Back. + MLP {(b)}
    & \xmark
    & \underline{47.5}
    & \bf 62.1
\\
Frozen Back. + MLP {(b)}
    & \cmark
    & 36.9 
    & 58.7 
\\
Full finetuning {(c)}
    & \xmark
    & \bf 50.5  
    & 58.9
\\
Joint distill. \& classif. as in \cite{peng2025learning}
    & \xmark
    & 44.9 
    & 56.0
\\
\bottomrule
\end{tabular}
\end{center}
\caption{\textbf{Downstream training recipe} using WI-768 backbone.}
\label{tab:experiments_what_to_train}
}
\end{table}

\paragraph{Effect of intensity.} 
We notice in \cref{tab:experiments_uni_modal_ingredients} that using intensity as input feature is detrimental for MUNet with both batchnorms or layernorms, and for WI-256 with batchnorms. This result is in line with the conclusions drawn in~\cite{Sanchez_2023_ICCV}. For WaffleIron with layernorms, the outcome is dependent on the setting: removing the intensity leads to a large gain on N$\shortto$K, a drop on N$\shortto$W, and no effect in the two other settings. The average mIoU over all dataset pairs is nevertheless better when intensity is not used. In the rest of the paper, we do not use intensity as input feature as this is the most robust strategy overall. 

\paragraph{Effect of normalization layers.} 
Due to sensor differences, the distribution of points in each dataset widely differs. It impacts feature distribution when going from the source to the target dataset, altering target segmentation performance. When using batch normalization layers, one way to address these changes, at least partially, is to adapt the batchnorm statistics to the target dataset~\cite{LI2018109, nado2020evaluating}. We study here an alternative: replacing all batchnorms (BNs) with layernorms (LNs). Averaging the mIoUs over all dataset pairs in \cref{tab:experiments_uni_modal_ingredients}, we notice that layernorms improve the performance for both backbones considered: +1.4 mIoU pts for MUNet, +5.3 for WI-256. Unlike batchnorms that use training feature statistics at inference, layernorms center and normalize the features using the actually seen statistics at inference. To maximize performance, we use layernorms in the rest of the paper.

\paragraph{Effect of backbone and capacity.}
WI-256 performs better than MUNet in all four pairs of datasets (\cref{tab:experiments_uni_modal_ingredients}).
For WaffleIron, we study the effect of increasing its feature size from 256 to 768. While increasing the number of parameters could have led to overfitting to the source dataset, we notice (\cref{tab:experiments_uni_modal_ingredients}) that WI-768 actually generalizes better and surpasses WI-256. 

Interestingly, WI-768 trained on source without adaptation outperforms the current SOTA in two settings out of four. 
This advocates for more studies in UDA on architectural choices.

%
\subsection{Choice of a visual foundation model for pretraining} 
\label{ssec:whic_choose}
For the remaining part of our study, the backbones are pretrained by distilling a VFM. The choice of VFM is key to get a good performance for downstream semantic segmentation. Several VFMs are studied in \cite{puy2024three} and the best results are obtained with DINOv2~\cite{oquab2023dinov2}. SAM~\cite{zou2023segment} is another powerful model also leveraged in, e.g., \cite{peng2025learning} for robust point cloud segmentation across domains. We test whether SAM is a better choice.

\paragraph{Pretraining protocol.} 
The lidar backbones are pretrained using ScaLR~\cite{puy2024three}. 
This VFM-based pretraining method requires calibrated and synchronized cameras and lidars to establish correspondences between points and pixels.
It does not need manual annotations. 
The training loss, optimization parameters and augmentations are described in the supp.\,mat.\ A.

\paragraph{SAM vs DINOv2.} 
We distill the knowledge from SAM \mbox{ViT-L} and DINOv2 \mbox{ViT-L} into WI-256 on nuScenes and SemanticKITTI jointly (the datasets are merged). Then, we freeze the weights of the pretrained backbone and train an MLP classification head on nuScenes and test the performance of the overall network on SemanticKITTI (and vice versa). We present the results in \cref{tab:experiments_sam_vs_dino}. We notice a better performance after distillation of DINOv2, showing that the features of DINOv2 are more suited for semantic understanding than those of SAM. For images, let us mention that a study of the properties of DINOv2 and SAM features was, e.g., conducted in \cite{ranzinger2024radio}. The results showed that SAM performs worse than DINOv2 for ``high-level object description'' and for ``combining the semantics of multiple objects.''. Our results indicate that this conclusion remains valid after distillation into lidar backbones.

%
\subsection{Downstream training}

After pretraining, the downstream training recipe is also important to get competitive results. We should make sure to exploit at best the features distilled from the chosen VFM and avoid scenarios where the backbone degenerates to the source-only performance.

\paragraph{Downstream training recipes.} 
We study three different ways for downstream training on semantic segmentation. 
(a)~We keep the lidar backbone weights frozen and train a linear classification head, with a batchnorm followed directly by a linear layer (note that these two layers can be combined into a single linear layer at inference). 
(b)~We keep the lidar backbone weights frozen and train a 2-layer MLP with ReLU activation.
(c)~We finetune the backbone weights along with the linear classification head (full finetuning).
Implementation details, such as epoch number and learning rates, are described in the supplementary material A.

\paragraph{MLP classification head performs the best.}
We present in \cref{tab:experiments_what_to_train} the results obtained with the three considered recipes. First, we notice that training only a linear classification head is underperforming compared to the other alternatives. 
Second, training an MLP classification head leads to similar results on average than finetuning the whole backbone. 
Yet, training an MLP is simpler and faster: full finetuning needs a careful choice of the learning rates applied on the pretrained backbones and on the classification head, and requires a full backward pass in the backbone. 
We also notice that the performance on the target dataset, after a first phase of improvement, keeps decreasing with full finetuning.
Instead, when using a simple MLP head, the performance stabilizes both on the source and target datasets.
This makes it an easy and practical method to use, as one just need to monitor the performance on the source dataset to stop the training.

\paragraph{Effect of intensity when pretraining.} 
Again in \cref{tab:experiments_what_to_train}, one can notice that the use of the intensity as input feature remains detrimental, even when pretraining on the source and target datasets and training a simple MLP head. This result stays in line with our design choice made in \cref{sec:arch}, {to disregard intensity when having to change domain}.

\begin{table}
\parbox{.48\linewidth}{
\scriptsize
\setlength{\tabcolsep}{3pt}
\begin{center}
\begin{tabular}{lcccc|cccc}
\toprule
\multirow{2}{*}{\shortstack{\\Back-\\bone}}
    & \multicolumn{3}{c}{Pretrain}
    & \multirow{2}{*}{\shortstack{Frozen\\back-\\bone}}
    & \multirow{2}{*}{\scriptsize N$\shortto$K}
    & \multirow{2}{*}{\scriptsize K$\shortto$N}
    & \multirow{2}{*}{\scriptsize N$\shortto$W}
    & \multirow{2}{*}{\scriptsize W$\shortto$N}
\\
\cmidrule{2-4}
    & S
    & T
    & E
    & \multicolumn{1}{c|}{}
\\
\midrule
\multirow{4}{*}{{WI-256}}
    & \xmark
    & \xmark
    & \xmark
    & \xmark
    & 41.1
    & 49.8
    & 35.3 
    & 61.8 
\\
    & \cmark
    & \xmark
    & \xmark
    & \cmark 
    & 31.0
    & 39.0
    & 30.2 
    & 32.8 
\\
    & \cmark
    & \cmark
    & \xmark
    & \cmark
    & 46.6
    & 59.7
    & 61.3 
    & 59.7 
\\
\midrule
\multirow{3}{*}{{WI-768}}
    & \xmark
    & \xmark
    & \xmark
    & \xmark
    & 44.6
    & 55.1
    & 37.1
    & 64.6
\\
    & \cmark
    & \cmark
    & \xmark
    & \cmark
    & 47.5
    & 62.1
    & 66.1
    & 70.5
\\
    & \cmark
    & \cmark
    & \cmark
    & \cmark
    & \bf 52.1 
    & \bf 63.4 
    & \bf 67.1 
    & \bf 71.8
\\
\midrule
\rowcolor{gray!20}
\multicolumn{9}{l}{\it Results of ScaLR \cite{puy2024three} with intensity. $^*$w/o Waymo.}
\\
\multirow{3}{*}{{WI-768}}
    & \xmark
    & \xmark
    & \xmark
    & \xmark
    & 28.6
    & -
    & -
    & -
\\
    & \cmark
    & \xmark
    & \xmark
    & \xmark
    & 29.6
    & -
    & -
    & -
\\
    & \cmark
    & \cmark
    & \cmark$^*$
    & \xmark
    & 36.6
    & -
    & -
    & -
\\
\bottomrule
\end{tabular}
\end{center}
\caption{\textbf{Adaptation performance} depending on the backbone, pretraining datasets (S, T, E), and {backbone freezing status}.
}
\label{tab:experiments_how_to_distill}
}
\hfill
\parbox{.48\linewidth}{
\scriptsize
\setlength{\tabcolsep}{3pt}
\begin{center}
\begin{tabular}{lccc|cccc}
\toprule
    \multirow{2}{*}{\shortstack{\\Back-\\bone}}
    & \multicolumn{2}{c}{Pretrain}
    & \multirow{2}{*}{\shortstack{\\Self-\\train}}
    & \multirow{2}{*}{\scriptsize N$\shortto$K}
    & \multirow{2}{*}{\scriptsize K$\shortto$N}
    & \multirow{2}{*}{\scriptsize N$\shortto$W}
    & \multirow{2}{*}{\scriptsize W$\shortto$N}
\\
\cmidrule{2-3}
    
    & S+T
    & S+T+E
    & \multicolumn{1}{c|}{}
\\
\midrule
\multirow{6}{*}{{WI-768}}
    & -    
    & -  
    & \xmark  
    & 44.6 
    & 55.1 
    & 37.1  
    & 64.6  
\\
    & -      
    & -  
    & \cmark
    & 45.7   
    & 55.6
    & 39.0 
    & 62.9 
\\
    & \cmark       
    & -  
    & \xmark
    & 47.5   
    & 62.1 
    & 66.1 
    & 70.5 
\\
    & \cmark        
    & -
    & \cmark
    & 49.5 
    & 66.5 
    & 69.8 
    & 69.6 
\\
    & -
    & \cmark
    & \xmark
    & 52.1
    & 63.4 
    & 67.1 
    & \bf 71.8 
\\
    & - 
    & \cmark 
    & \cmark 
    & \bf 52.1 
    & \bf 66.4
    & \bf 69.1 
    & 70.5 
\\
\bottomrule
\end{tabular}
\end{center}
\caption{\textbf{Effect of self-training}. We present the consistent benefit of self-training whether the backbone is not pretrained, pretrained on {both the source and target dataset (S+T), or pretrained on all the considered datasets (S+T+E).}}
\label{tab:self_training}
}
\end{table}

\paragraph{Joint distillation and semantic segmentation.} Instead of first pretraining the source and target datasets, and then finetuning the backbone for semantic segmentation on the source dataset, one can naturally wonder if optimizing for both tasks at the same time (distillation + semantic segmentation), as done in~\cite{peng2025learning}, is a better choice or not. 

We test this design as follows. Two different heads are added at the end of the lidar backbone: one for distillation and one for classification.
The backbone sees both source and target point clouds. The distillation loss is computed for both source and target point clouds. The classification loss (cross-entropy plus Lov\'asz loss) is applied on source point clouds. We present the result of this strategy in the last row of \cref{tab:experiments_what_to_train}. We observe that the strategy is the worst on K$\shortto$N and is the second worse on N$\shortto$K.

%
\subsection{Choice of pretraining datasets}
Finally, we study the choice of pretraining datasets. A few results in \cite{puy2024three} suggest that the generalization capabilities of the lidar backbone improves when combining multiple datasets for pretraining. In this section, we analyze this behavior much more thoroughly by testing exhaustively different dataset combinations. We specialize this study to the best setup we have constructed so far for UDA.

\begin{table}[t]

    \parbox{.65\linewidth}{
    \centering
    \scriptsize
    \setlength{\tabcolsep}{4.5pt}
    \begin{tabular}{llc|cccc|c}
    \toprule
    Method
        & Back.
        & Mod.
        & {N$\shortto$K}
        & {K$\shortto$N}
        & {N$\shortto$W}
        & {W$\shortto$N}
        & Avg.
    \\
    \midrule
    \rowcolor{gray!20}
        Target oracle
        & MUNet
        & - 
        & 70.3 & 78.3 & 79.9 & 78.3 & 76.7
    \\ 
    \rowcolor{gray!20}
        Target oracle 
        & WI-768
        & -
        & 72.4 & 83.8 & 83.4 & 83.8 & 80.9
    \\
    \midrule
        Source only$^\dagger$
        & MUNet
        & - 
        & 27.7 & 28.1 & 29.4 & 21.8 & 26.8
    \\
        Source only
        & WI-768
        & - 
        & 44.6 
        & 55.1 
        & 37.1
        & 64.6
        & 50.4
    \\
    \midrule
        PL$^\dagger$~\cite{morerio2017minimal}
        & MUNet
        & U 
        & 30.0 & 29.0 & 31.9 & 22.3 & 28.3
    \\
        CosMix$^\dagger$~\cite{saltori2022cosmix}
        & MUNet
        & U 
        & 30.6 & 29.7 & 31.5 & 30.0 & 30.5
    \\
        MM2D3D$^\dagger$~\cite{cardace2023exploiting} 
        & MUNet
        & M 
        & 30.4 & 31.9 & 31.3 & 33.5 & 31.8
    \\
        MM2D3D*$^\dagger$~\cite{cardace2023exploiting} 
        & MUNet
        & M 
        & 32.9 & 33.7 & 34.1 & 37.5 & 34.6
    \\
        Adapt-SAM$^\dagger$~\cite{peng2025learning} 
        & MUNet
        & M 
        & 48.5 & 42.9 & 44.9 & 48.2 & 46.1
        \\
        \rowcolor{orange!15}
        \method (ours)  & MUNet & M & 41.8&  53.3 & 54.1 & 53.0 & 50.6\\

    \midrule
        Self-Training
        & WI-768
        & U
        & 45.7   
        & 55.6
        & 27.6
        & 48.0
        & 44.2
    \\
    \rowcolor{orange!15}
        \method (ours)
        & WI-768
        & M 
        & \bf 52.1 & \bf 66.4 & \bf 69.1 & \bf 70.5 & \bf 64.5
    \\
    \bottomrule
    \end{tabular}
    
    \begin{flushleft} \footnotesize
        \textbf{(a) Comparison to existing approaches.} The results for methods marked with $^\dagger$ are reported from~\cite{peng2025learning}. {The current} absence of code {for \cite{peng2025learning}} prevents us from testing it with our architecture choices.
    \end{flushleft}
    }
    \hfill
    \parbox{.33\linewidth}{
    
        \scriptsize
        \setlength{\tabcolsep}{2pt}
        \centering
        \begin{tabular}{l|c|c}
            \toprule
            & \shortstack{\smash{Adapt-}\\SAM \cite{peng2025learning}} & \cellcolor{orange!15} \shortstack{MuDDoS\\(ours)} \\
            \midrule
             Distillation & Joint & \cellcolor{orange!15} Sequential\\
             Ext. Data  & \xmark & \cellcolor{orange!15} \cmark \\
             Finetuning & Full & \cellcolor{orange!15} Head\\
             VFM & SAM & \cellcolor{orange!15} DINOv2\\
             Backbone & MUNet & \cellcolor{orange!15} WI-768 \\
             Scaled backbone & \xmark & \cellcolor{orange!15} \cmark \\
             Intensity & \cmark & \cellcolor{orange!15} \xmark \\
             Norm & BN & \cellcolor{orange!15} LN \\ 
            \bottomrule
        \end{tabular}
        
        \begin{flushleft} 
            \textbf{(b) Main differences} with Adapt-SAM, previous SOTA.
        \end{flushleft}
    }
    
    \caption{\textbf{Comparison to the state of the art}. (a)~We compare  quantitatively our best setting (\method{}) to both unimodal (U) and multimodal (M) UDA methods for 3D semantic segmentation of lidar point clouds. (b)~We highlight the main changes introduced by our study with respect to the previous state of-the-art method, i.e., Adapt-SAM~\cite{peng2025learning}.
    }
    \label{tab:experiments_main_comparison}
    
\end{table}

\paragraph{Pretraining dataset configurations.}
We study three different settings (in \cref{tab:experiments_how_to_distill}).
\setlist{nolistsep}
\begin{itemize}[noitemsep]
    \item \textit{Source pretraining} (S): it corresponds to a source-only setting.
    \item \textit{Source \& Target pretraining} (S+T): pretraining is done on both source and target.
    \item \textit{Source \& Target \& External datasets pretraining} (S+T+E): pretraining is done combining nuScenes, SemanticKITTI and Waymo, to which we also add PandaSet~\cite{xiao2021pandaset}.
\end{itemize}
For each experiment, we pretrain the backbone using ScaLR, the weights of the backbone are then frozen, and the classification head (a 2-layer MLP) is trained using the source labels.

\paragraph{Impact of pretraining on both source and target.}

First, we observe that the results obtained for WI-256 with source pretraining and downstream training of the classification head are worse than those obtained when training WI-256 from scratch on source datasets.
The quality of the features after source pretraining is not good enough to reach a high performance on the target datasets using only an MLP classification head.
Nevertheless, as soon as the network is pretrained on both the source and target datasets, tuning the classification head on the source datasets is enough to surpass the results obtained in \cref{tab:experiments_uni_modal_ingredients} in 3 out of 4 settings. 
This demonstrate that pretraining on Source \& Target produces features: (a)~that are well aligned between the source and target domain, and (b)~that can be used as is by just training an MLP on top of them. 
As previously, we still observe better results with WI-768 vs WI-256.

\paragraph{Impact of pretraining on more datasets.}

Recent works~\cite{puy2024three,wu2024towards} highlight that training jointly with multiple datasets can improve feature quality. 
We show in \cref{tab:experiments_how_to_distill} that this observation also holds for domain adaptation. 
Indeed, the performance of WI-768 improves by at least 1.0 mIoU point and up to 4.6 points when pretrained on the mix of all datasets, compared to pretraining only on the source and target datasets. 

From a practical point of view, besides obtaining better performances, this result shows that one can pretrain a \textit{single backbone} to address \textit{multiple domain} shifts, rather than one pretraining for each pair of source and target datasets. In fact, it is possible to pretrain a single backbone that addresses all UDA settings together, and only train one classification head per source dataset. This {linear number of trainings contrasts with methods in which the network has to be retrained for each dataset pair, thus with a quadratic number of trainings. Besides, training only a 2-layer MLP in our case is more efficient than training a complete network.

We also remark that our results are much better than those in \cite{puy2024three} on N$\shortto$K despite similar pretrainings. It highlights the importance of our architectural choices (no use of intensity) and downstream training recipe (training a simple MLP head instead of full finetuning).

\subsection{Self-training for further improvements}

A classical recipe to improve results is to do a final round of self-training using a teacher-student mechanism. We complement our study by checking if this stage is also beneficial in our case. We train only the MLP head and keep the backbone frozen. We present results in \cref{tab:self_training}. We notice that it improves the performance in all cases for WI-768, except for W$\shortto$N where we observe a slight decrease. See supplementary material A for training details.

\clearpage

%
\section{Comparison to SOTA methods}

\begin{figure*}[t]
    \small
    \begin{center}
    \setlength{\tabcolsep}{6pt}
    \begin{tabular}{@{\hskip 3pt}l@{\hskip 5pt}c@{\hskip 5pt} c c c c}
        \multirow{2}{*}{{\rotatebox{90}{N$\shortto$K}}}&\rotatebox{90}{{\hskip 3pt} Source only}
        & \includegraphics[width=0.2\textwidth]{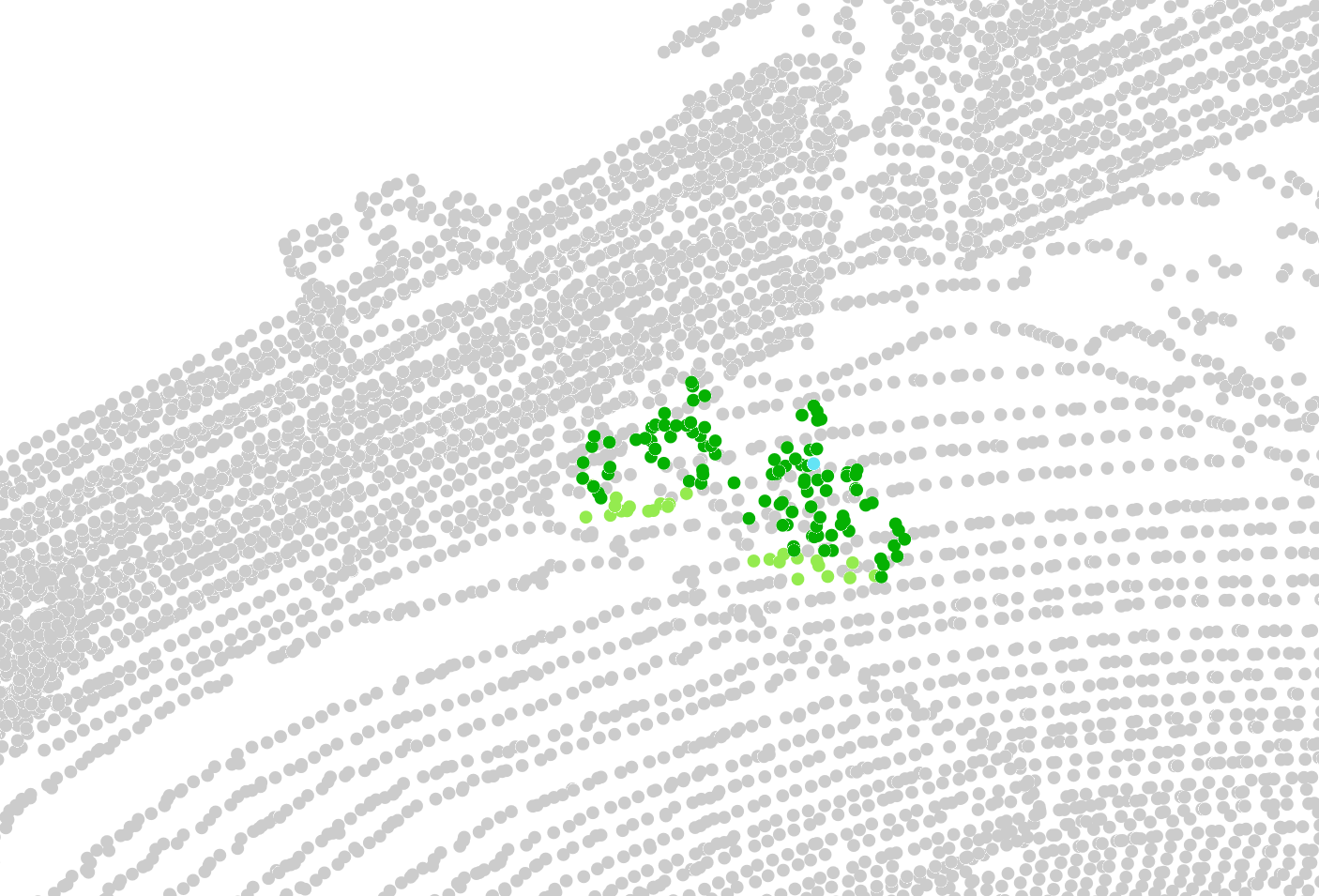} 
        & \includegraphics[width=0.2\textwidth]{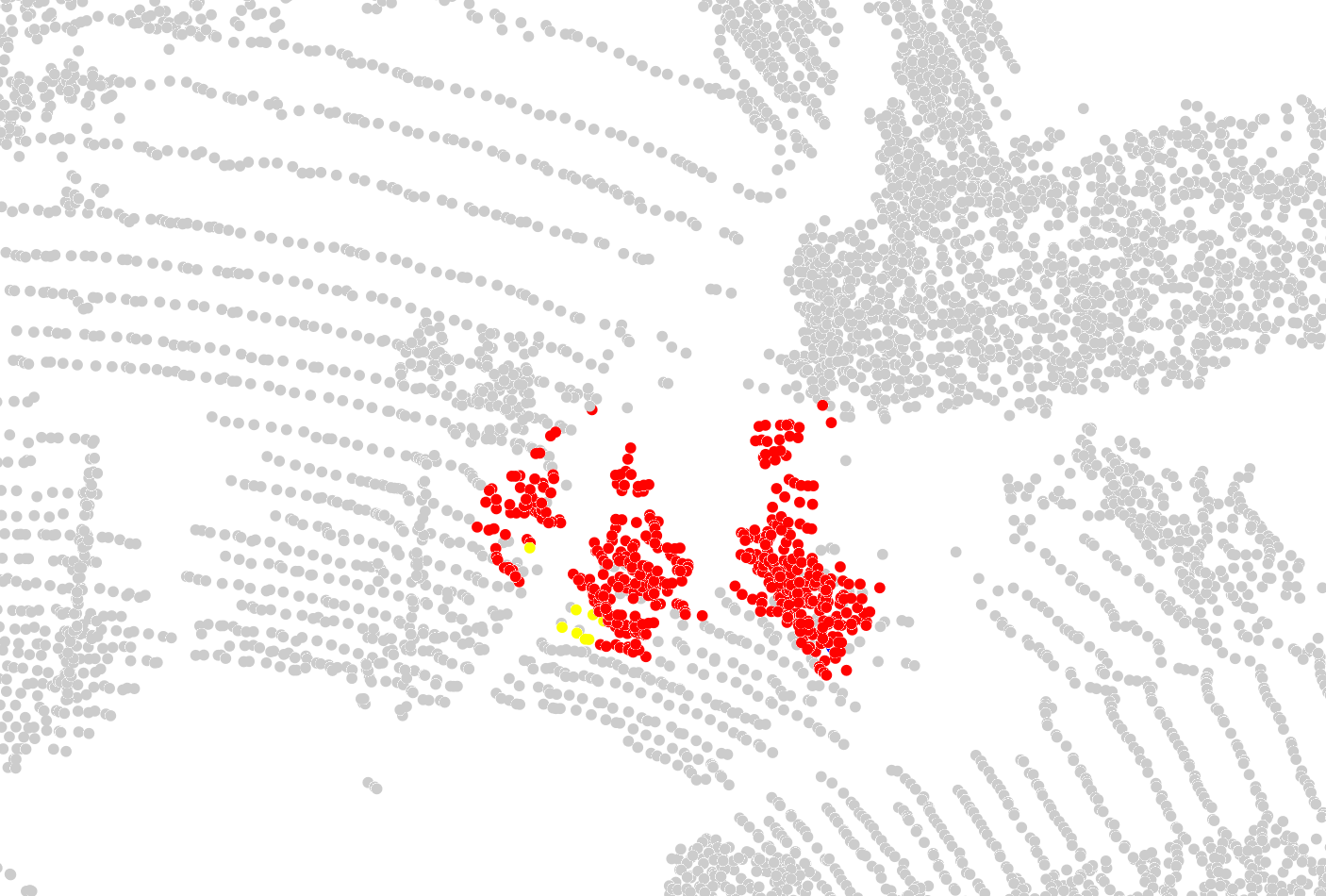} 
        & \includegraphics[width=0.2\textwidth]{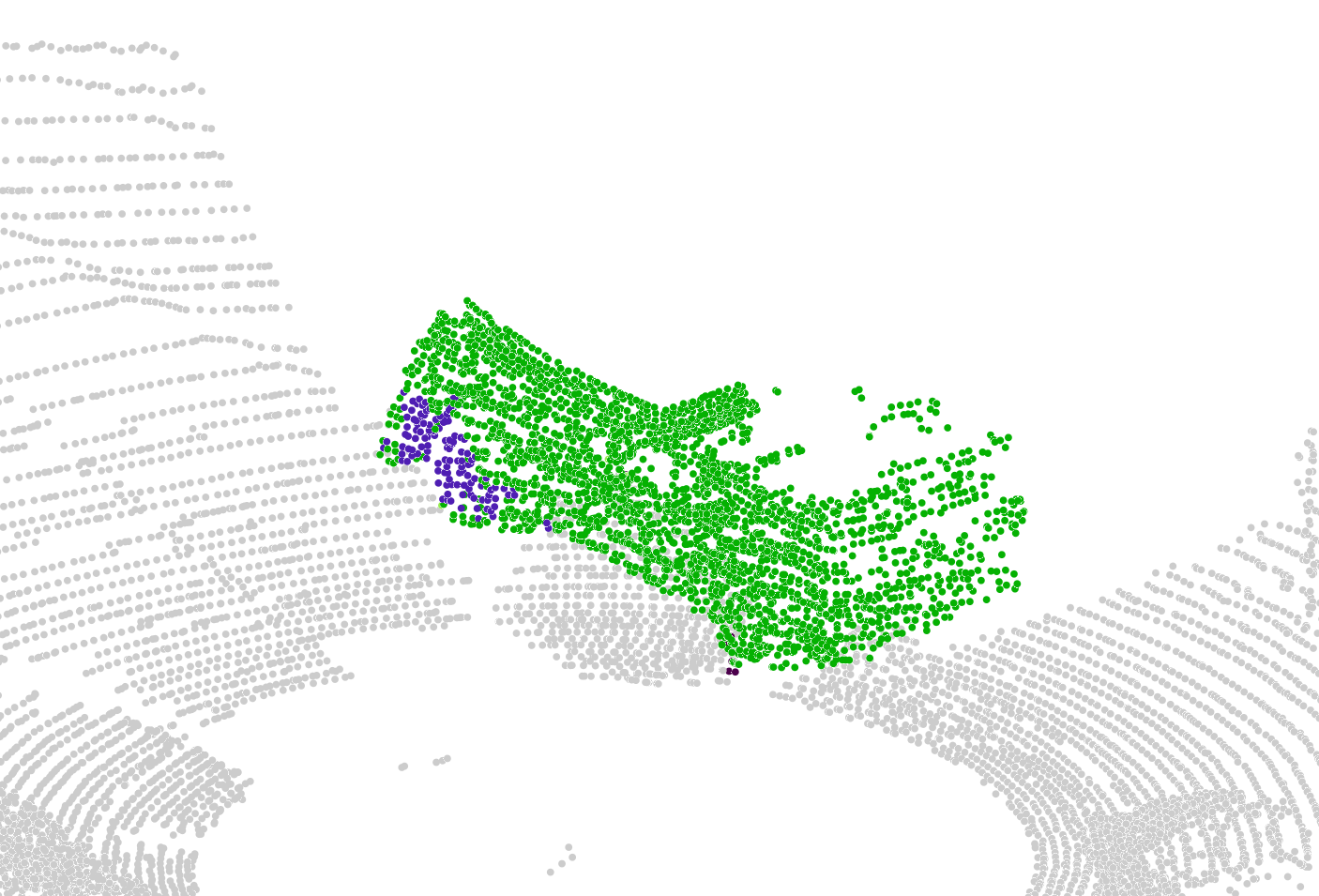} 
        & \includegraphics[width=0.2\textwidth]{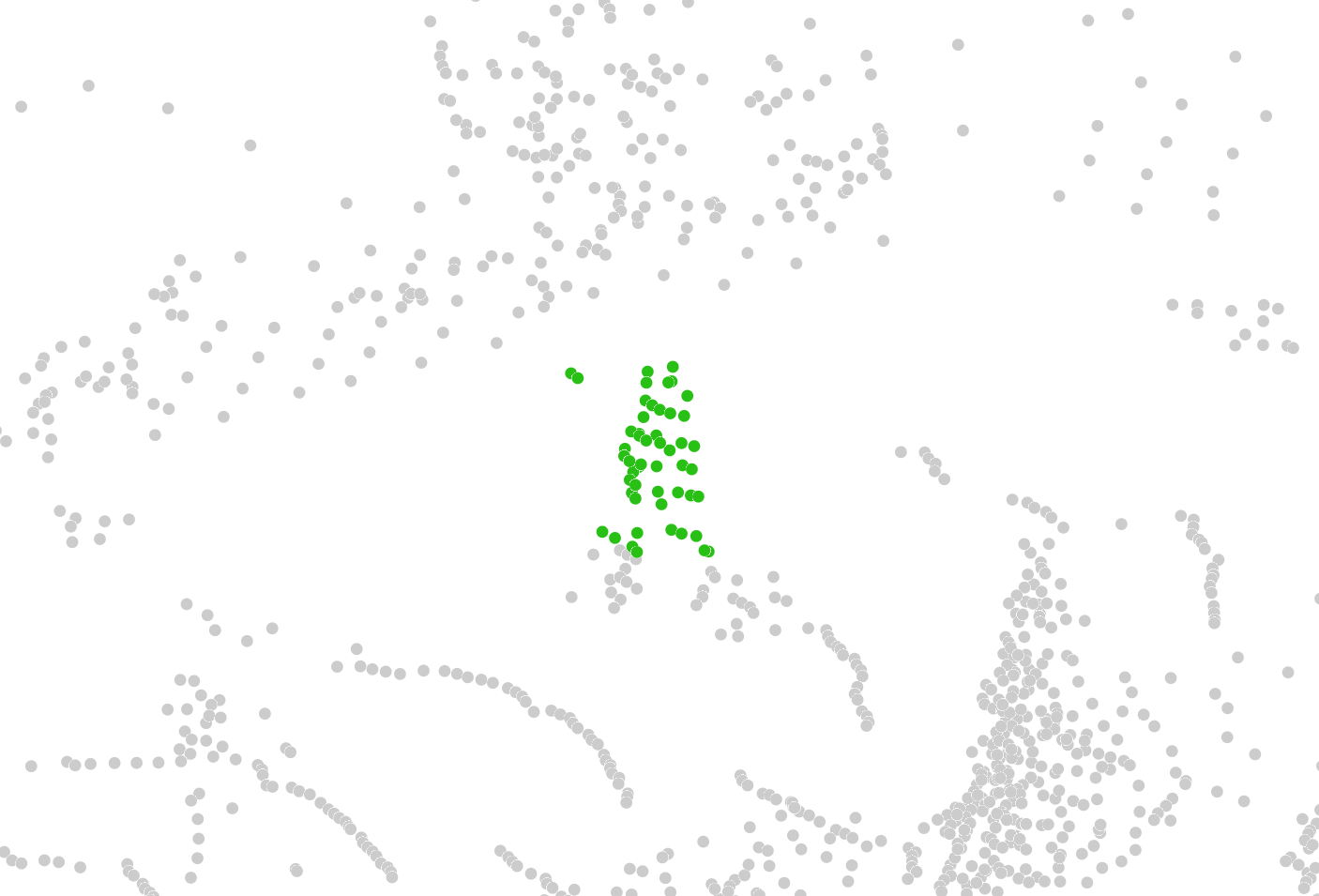} \\
        
        &\rotatebox{90}{{\hskip 5pt} \method}
        & \includegraphics[width=0.2\textwidth]{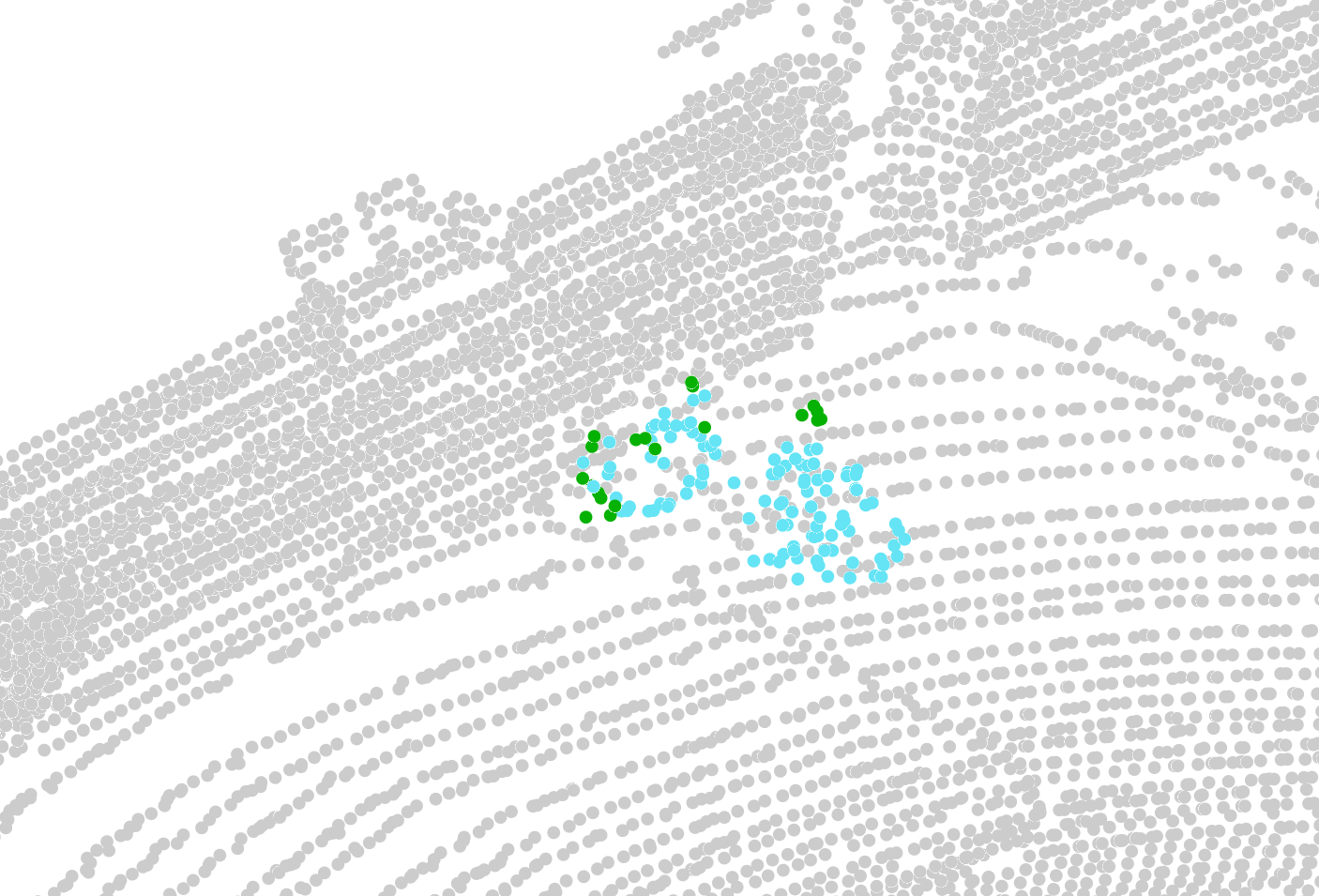} 
        & \includegraphics[width=0.2\textwidth]{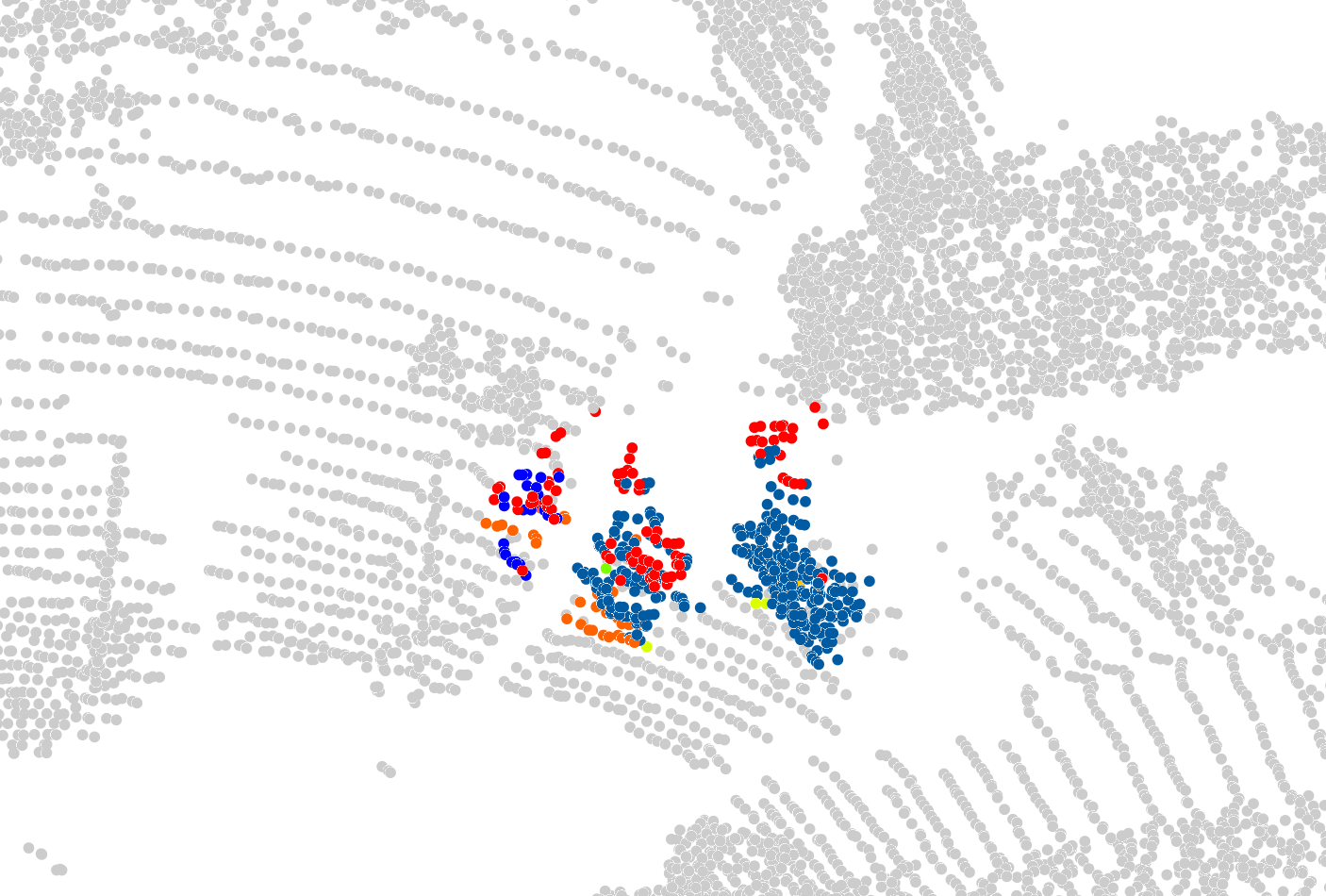} 
        & \includegraphics[width=0.2\textwidth]{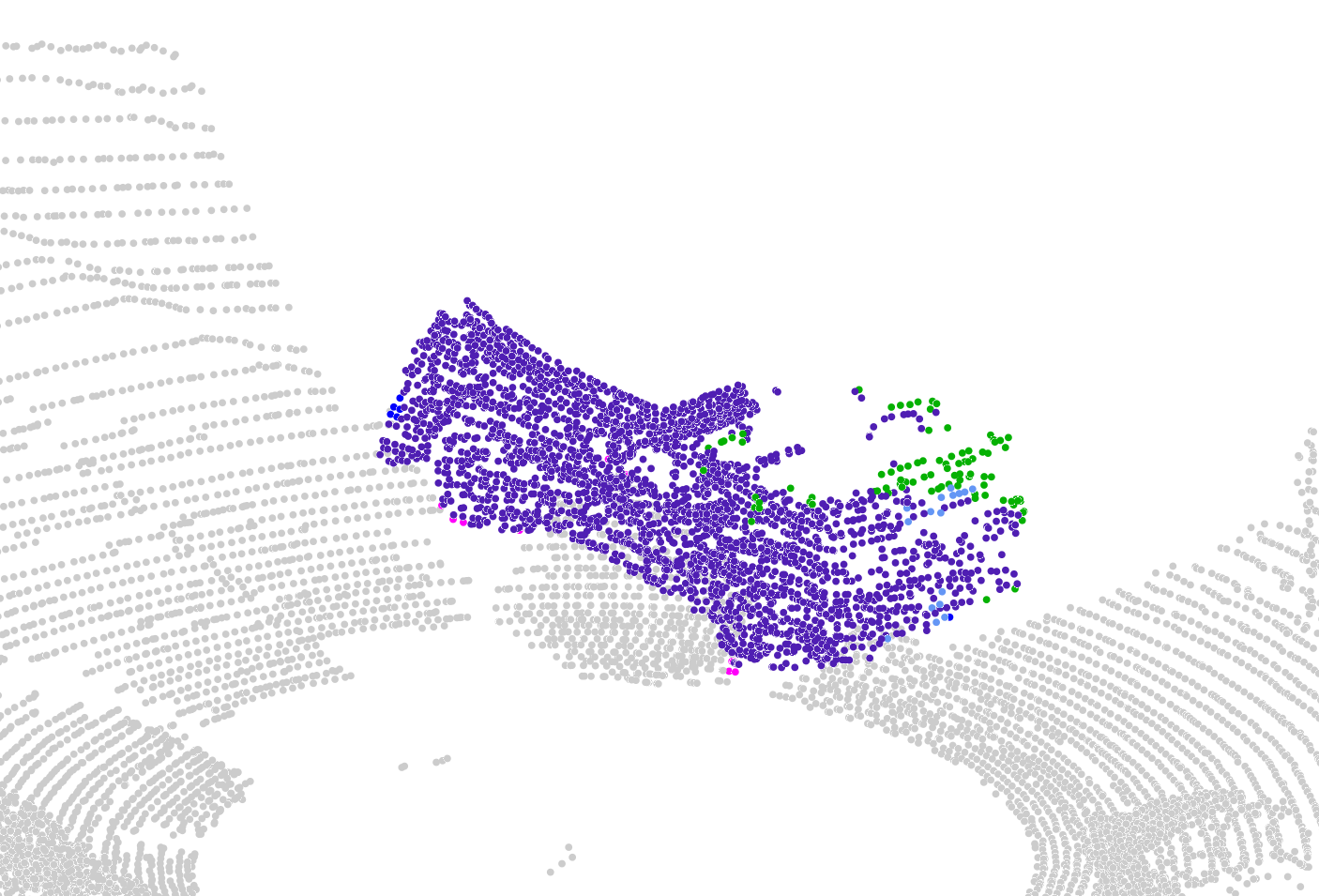} 
        & \includegraphics[width=0.2\textwidth]{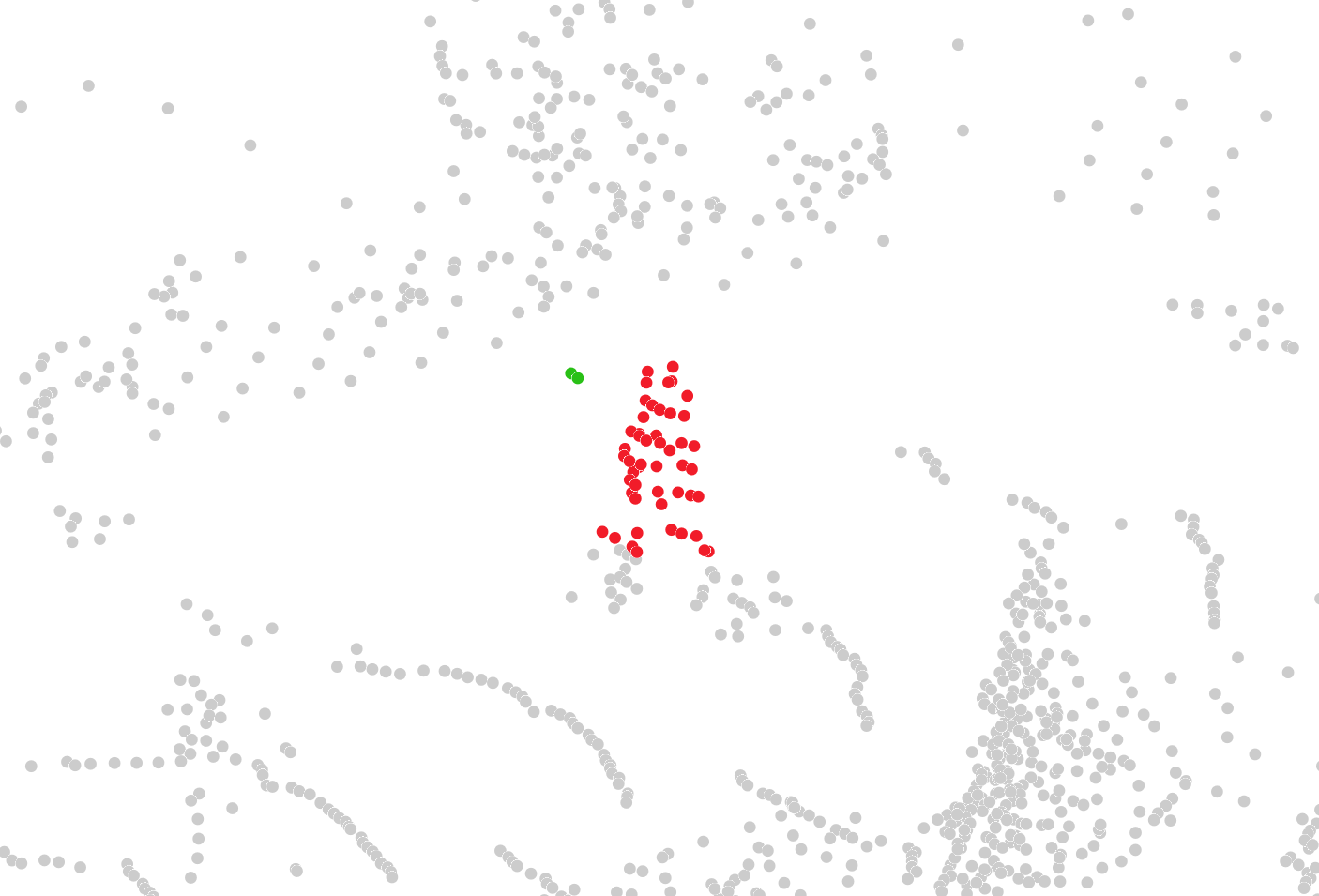} \\
        \midrule
        \multirow{2}{*}{{\rotatebox{90}{N$\shortto$W}}}&\rotatebox{90}{{\hskip 3pt} Source only}
        & \includegraphics[width=0.2\textwidth]{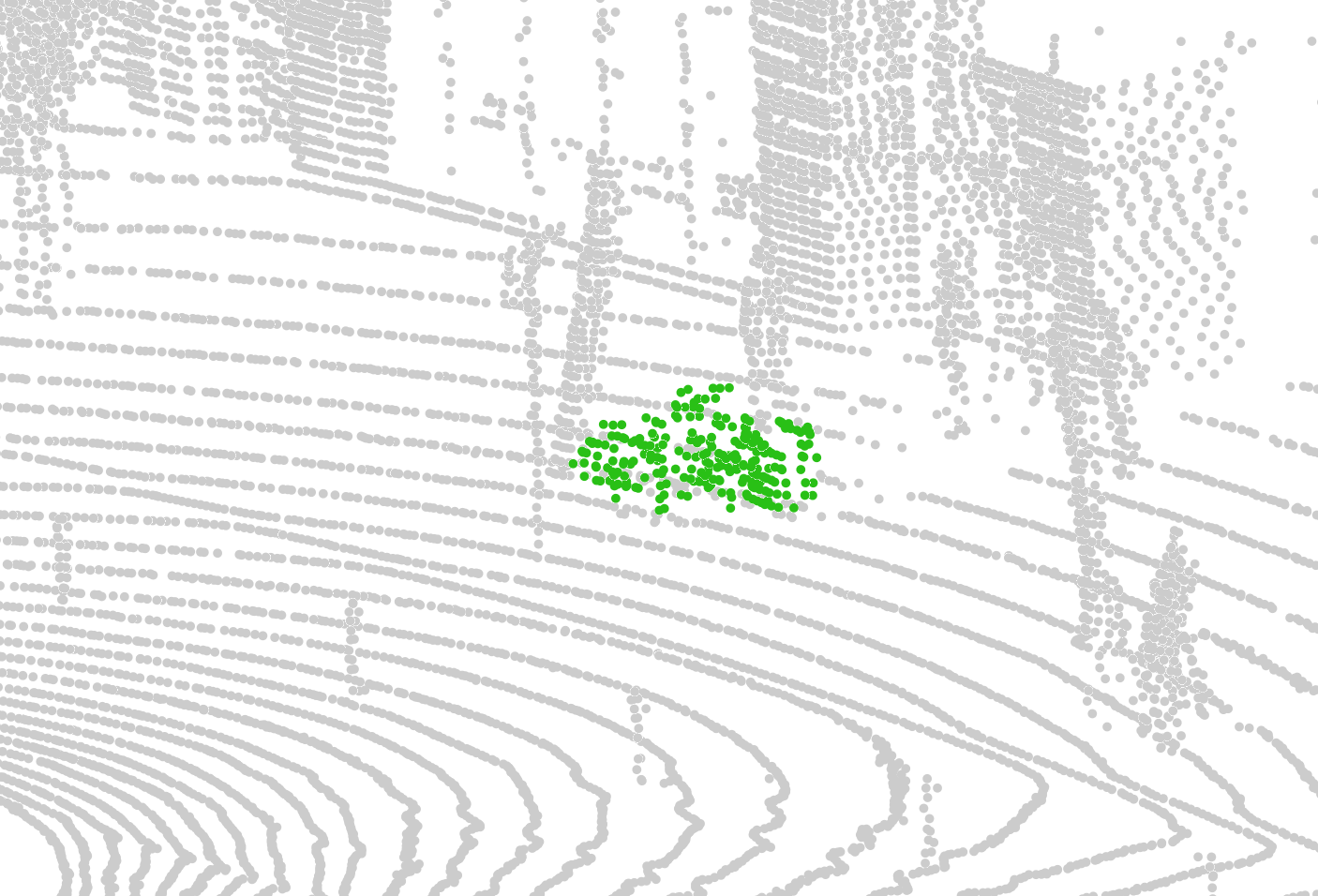} 
        & \includegraphics[width=0.2\textwidth]{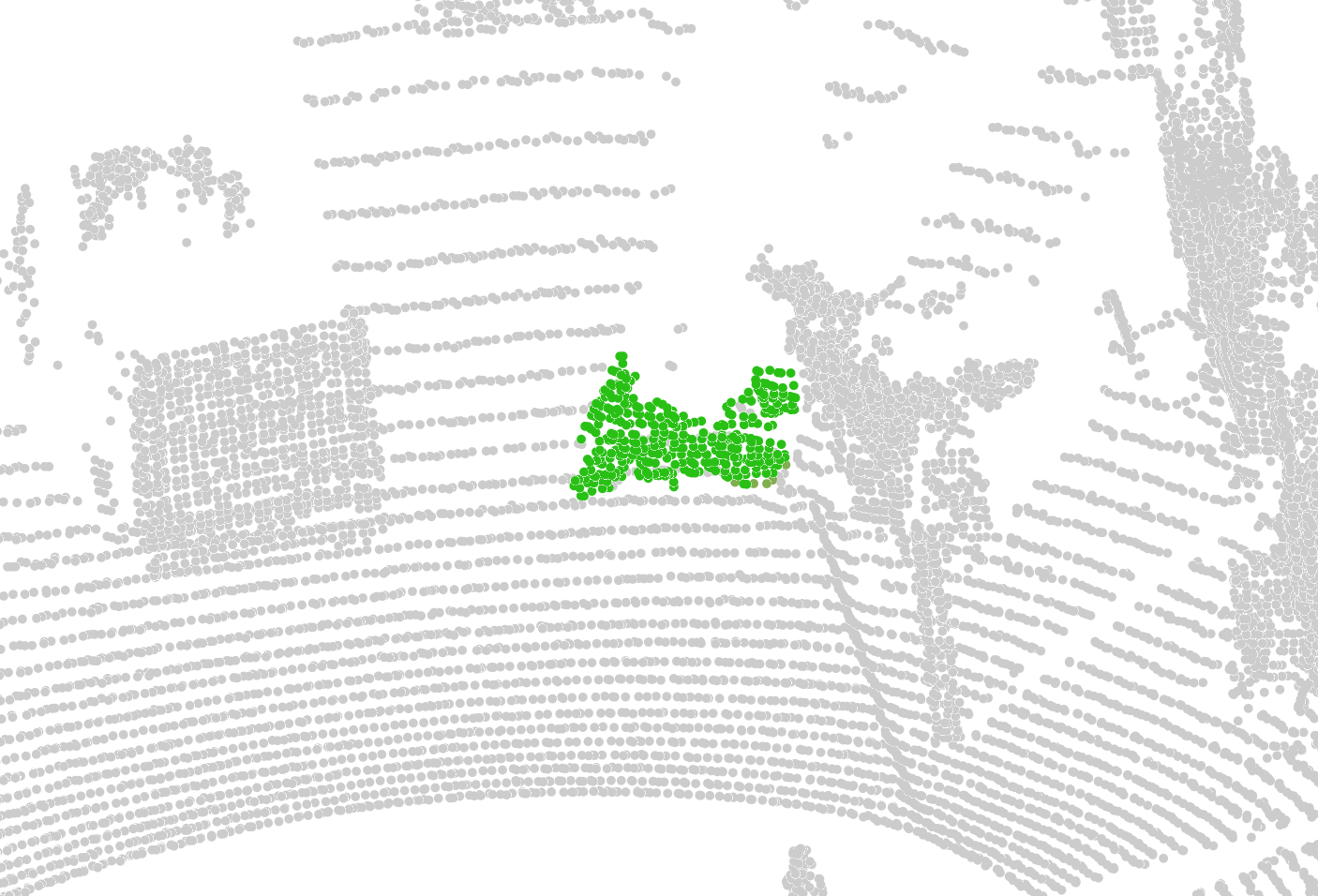} 
        & \includegraphics[width=0.2\textwidth]{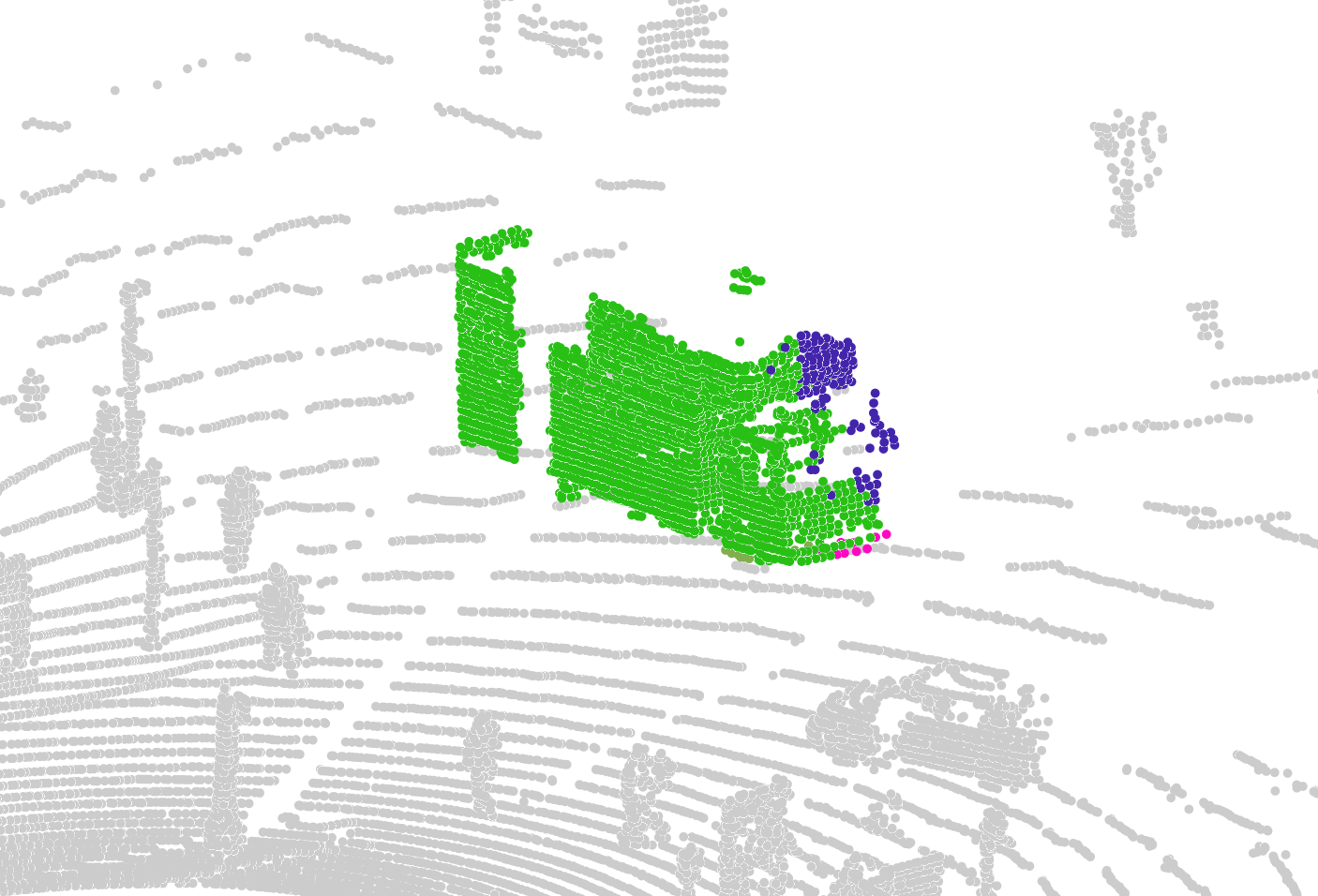} 
        & \includegraphics[width=0.2\textwidth]{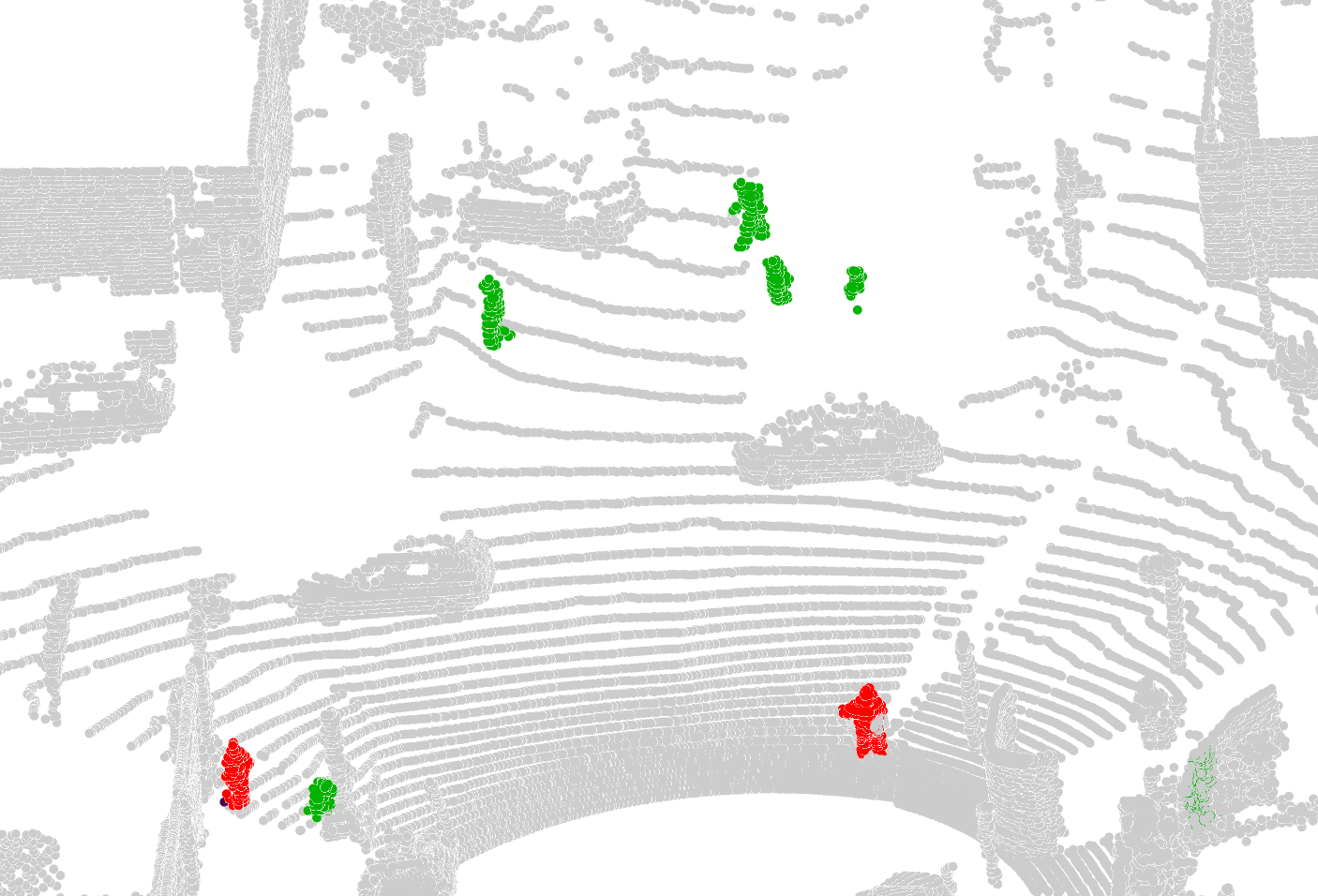} \\
        
        &\rotatebox{90}{{\hskip 5pt} \method}
        & \includegraphics[width=0.2\textwidth]{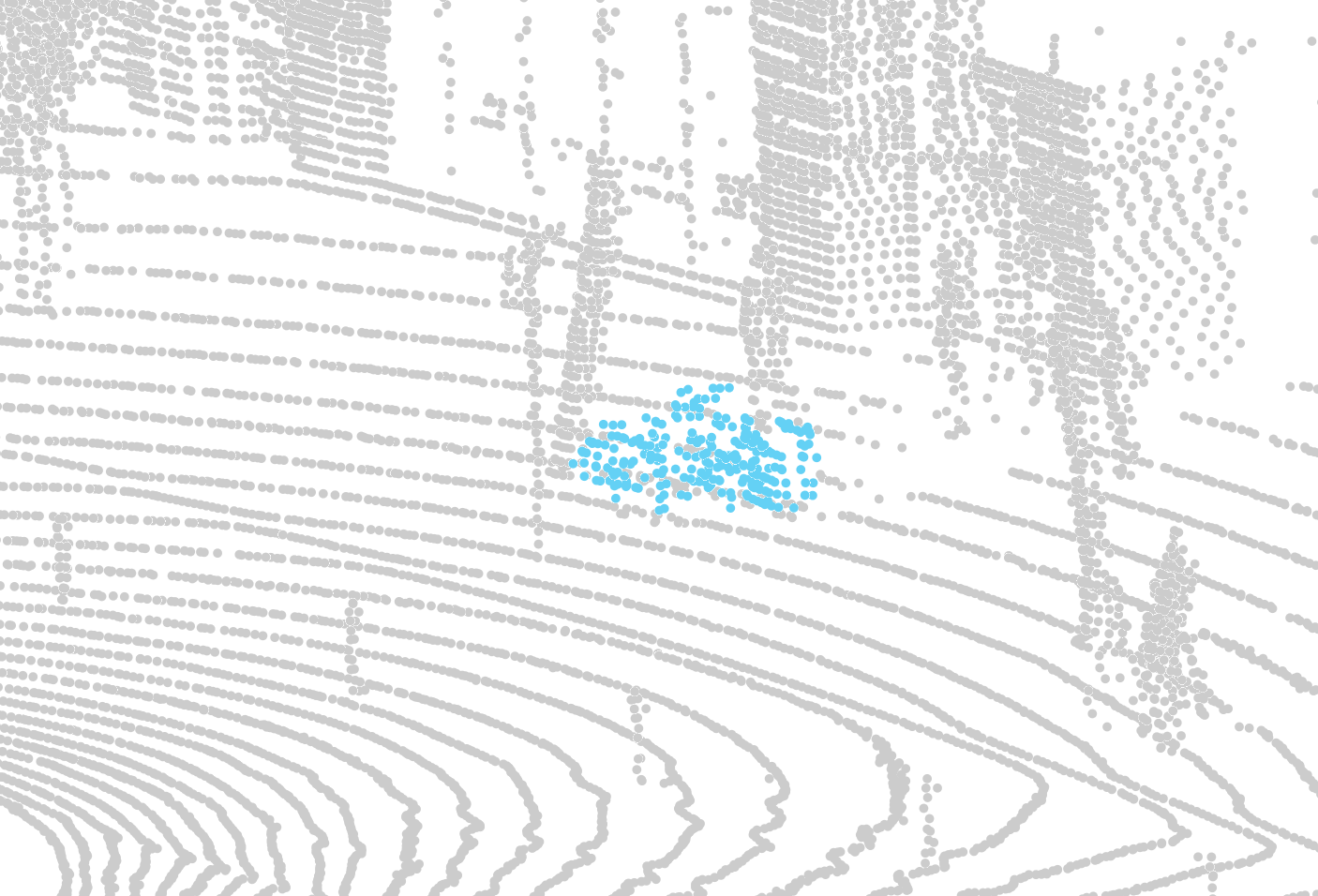} 
        & \includegraphics[width=0.2\textwidth]{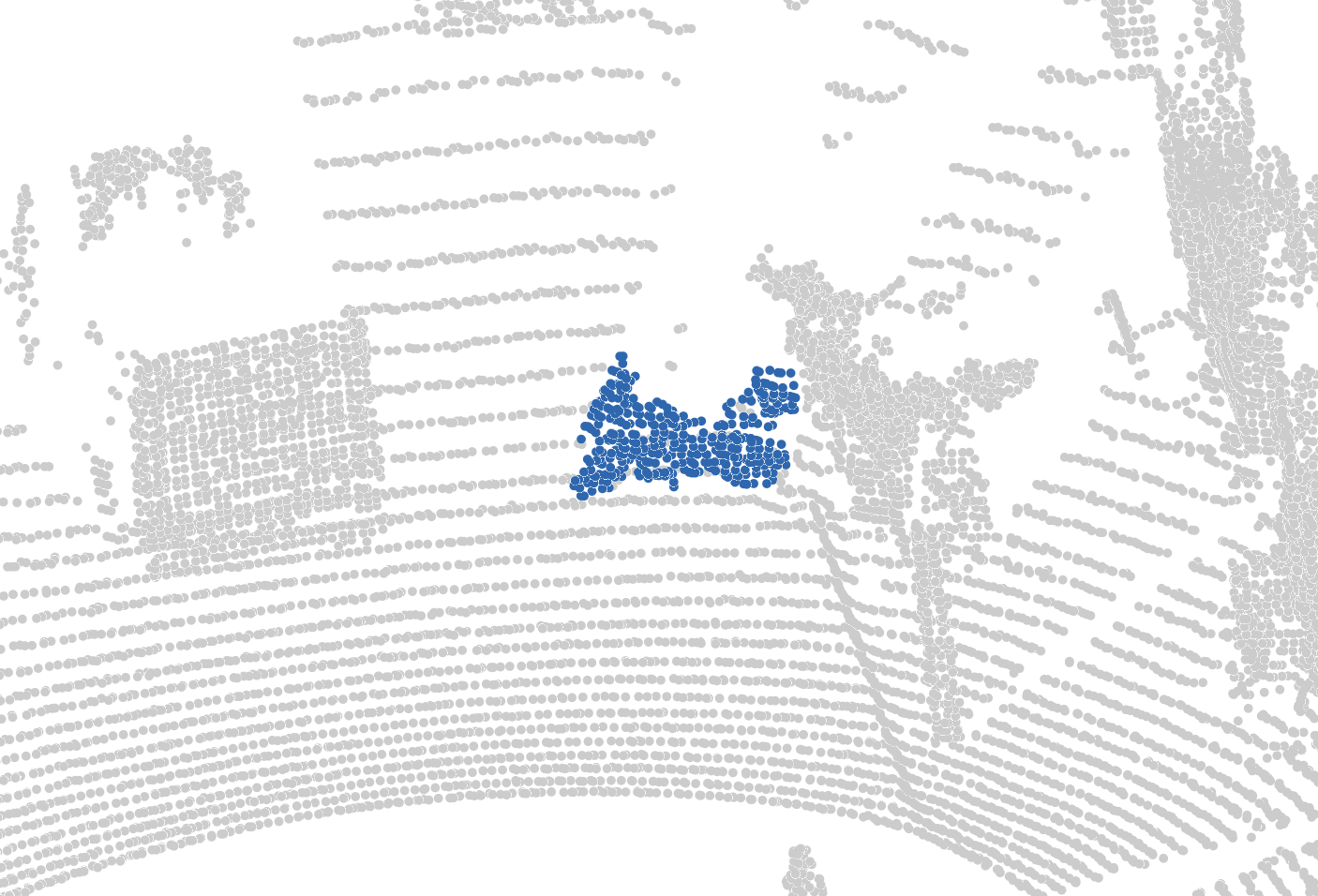} 
        & \includegraphics[width=0.2\textwidth]{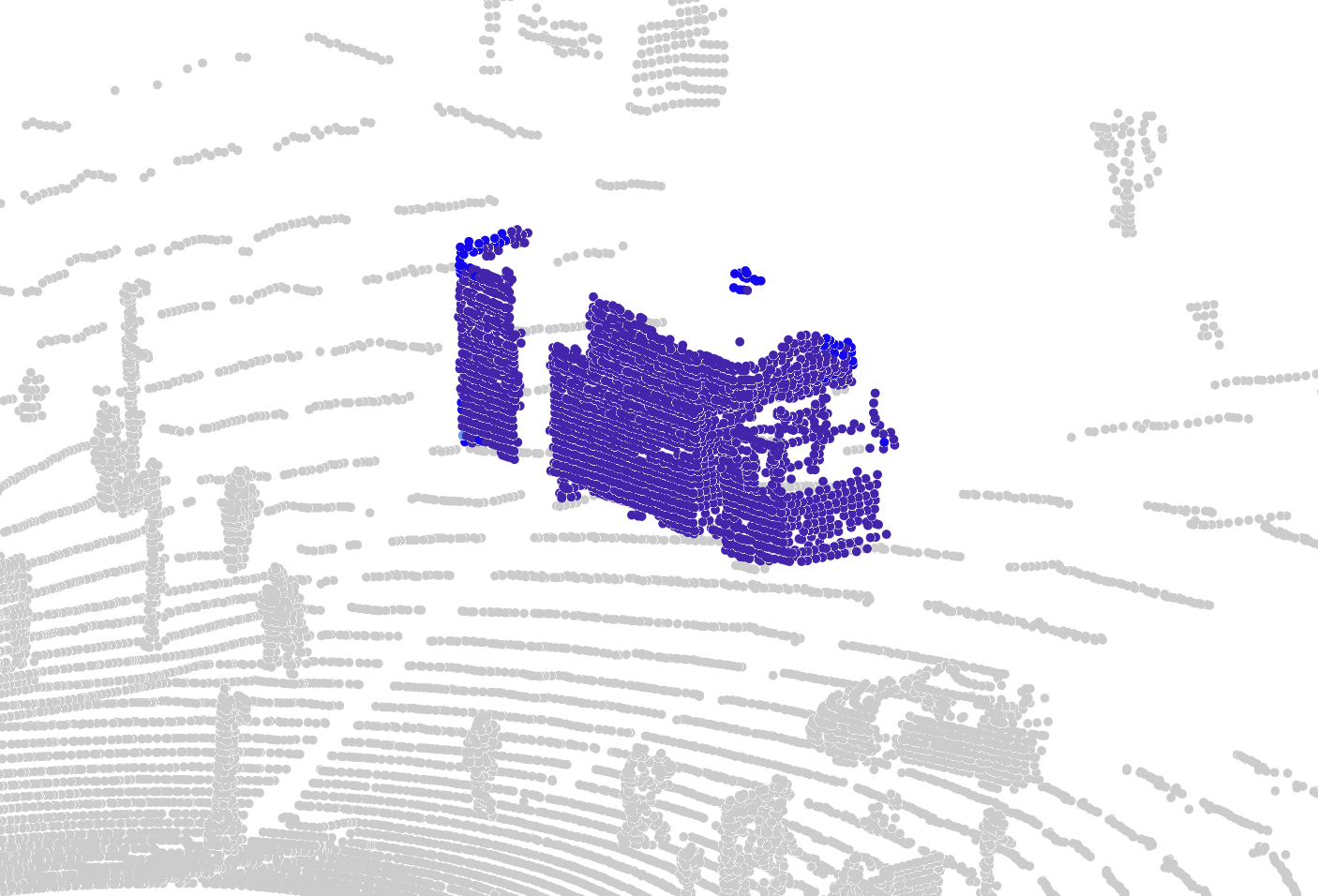} 
        & \includegraphics[width=0.2\textwidth]{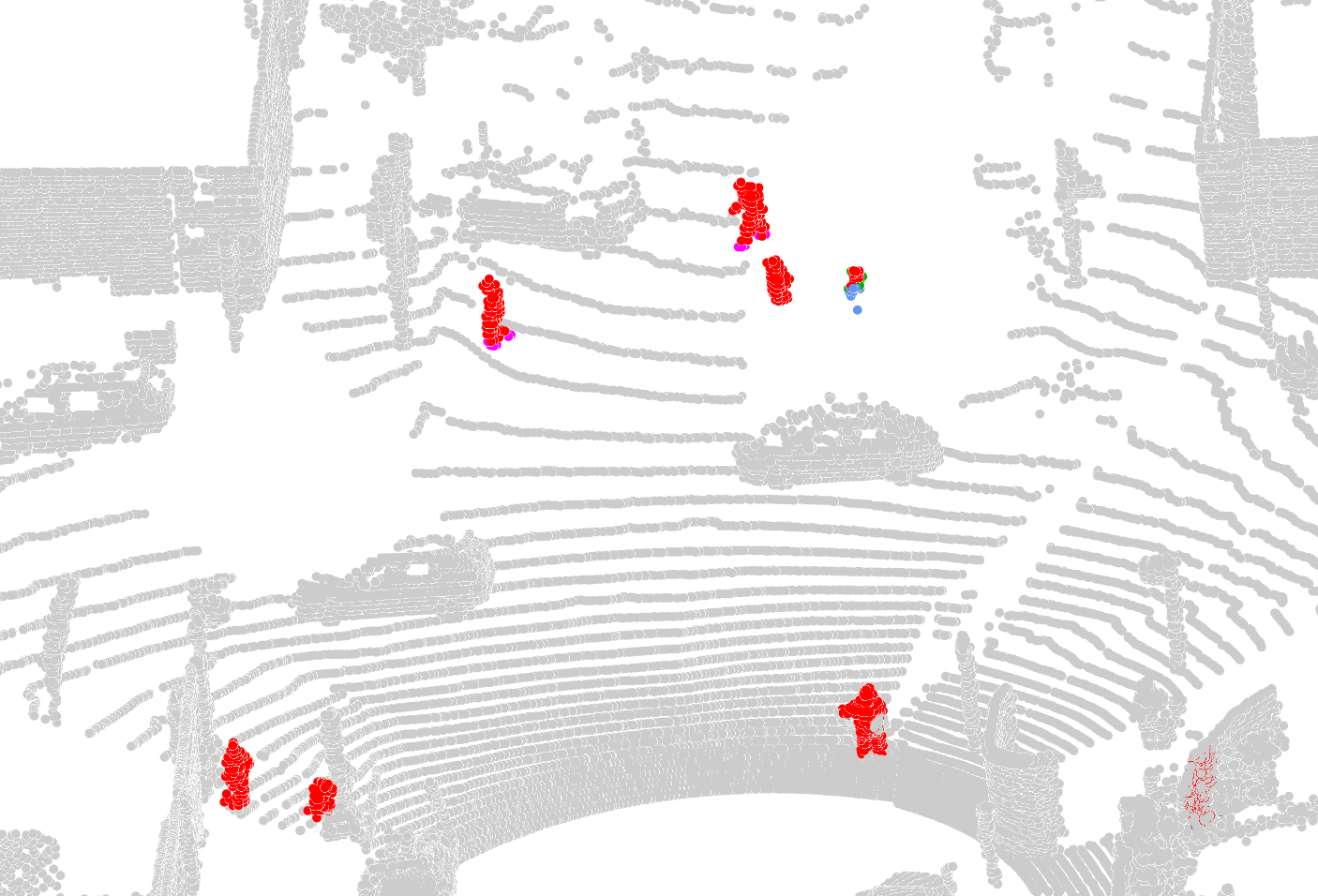} \\
        & & \colorbox{skbicycle}{\color{white}`Bicycle'}& \colorbox{skmotorcycle}{\color{white}`Motorcycle'}  & \colorbox{sktruck}{\color{white}`Truck' } & \colorbox{skpedestrian}{\color{white}`Pedestrian'} \\
        
    \end{tabular}
    
    \end{center}
    \caption{\textbf{Qualitative results on N$\shortto$K (Top) and N$\shortto$W (Bottom).} The label colors correspond to ground truth label assigned color. Points with a ground-truth not belonging to the shown class are grayed out. The source only model tends to over predict vegetation and sometimes mistakes dense partially occluded object with other classes, e.g., pedestrian instead of motorcycle in the second example. \method is able to partially or completely recover the correct classes.}
    \label{fig:qualitative}
\end{figure*}

In this section, we compare our method to state-of-the-art methods in unsupervised domain adaptation across lidars. We recall the final setup for our method.
\begin{itemize}
    \item \textit{Backbone}. We use WI-768 or MinkUNet as in \cite{saltori2022cosmix,peng2025learning}. In both cases, we exploit the findings in our study: we use layernorms and remove intensity from the input.
    \item \textit{Pretraining datasets}. The distillation is done on the combination of nuScenes, SemanticKITTI, Waymo and Pandaset (N\&K\&W\&P). The \textit{same} pretrained backbone is re-used for all cross-domain settings.
    \item \textit{Downstream training}. After distillation, the backbone weights are frozen and we train an MLP classification head on the source dataset.
\end{itemize}

\paragraph{High-level comparisons.} We present the results in \cref{tab:experiments_main_comparison}. First, we remark that our technique with MinkUNet performs better than Adapt-SAM with significant margin on three out of the four pairs of dataset considered. We also notice that our best discovered setup with WI-768 outperforms former results with very significant margins (up to 24.2 mIoU pts on N$\shortto$W, and 18.4 mIoU pts on average). It shows the interest of optimizing each stage of the training pipeline from the choice of the backbone to the downstream training recipes, by way of the pretraining datasets. 

In~\cref{tab:experiments_main_comparison}\textcolor{red}{(b)}, we contrast directly which of our findings differ from what has been used in the current SOTA method Adapt-SAM~\cite{peng2025learning}. At the time of submission, no code is available for~\cite{peng2025learning} to integrate our findings in their method.

\paragraph{Per class comparisons.}
The main comparisons in \cref{tab:experiments_main_comparison} are extracted from \cite{peng2025learning}, which, unfortunately, does not provide per class results. To get hints about the classes which perform well or not with our method, we compare it to results reported in~\cite{michele2024saluda}. Please refer to~\cref{tab:app:experiments_per_class_ns_sk_full} in the supplementary material for numbers. None of the methods in this comparison uses intensity as input features. We notice that easy classes such as car, vegetation and drivable surface perform well with all methods. The most notable differences (also to the source-only models) are on bicycle, motorcycle, truck and pedestrian, where our approach performs significantly better.

\paragraph{Qualitative results.}
We present in \cref{fig:qualitative} qualitative results on challenging classes. While not perfect, our method is able to correctly segment part of objects that are completely mislabeled with the source-only model.

\section{Conclusion}

The main messages of this work are that: (1)~the choice of the backbone architecture is a key part in the design of domain adaptation methods: simple changes can significantly improve the performance; (2)~it is possible to pretrain a single backbone to address many domain shifts thanks to the distillation of VFM features; (3)~the downstream training recipe should be designed with care to avoid degrading the quality of the pretrained features.

Looking ahead, this study opens several interesting perspectives.
Indeed, as the distillation process does not require any annotation, scaling the backbone and the size of the pretraining dataset even more could lead to improved generalization capabilities, offering foundation models for lidar point clouds that are as powerful as for images.
Future research directions include combining several VFMs, e.g., to benefit from the quality of DINOv2 features for semantic understanding and of SAM features for instance augmentations.

\section*{Acknowledgements}

We also acknowledge the support of the French Agence Nationale de la Recherche
(ANR), under grants ANR-21-CE23-0032 (project MultiTrans), ANR-20-CHIA-
0030 (OTTOPIA AI chair), and the European Lighthouse on Secure and Safe AI
funded by the European Union under grant agreement No. 101070617. This work
was performed using HPC resources from GENCI–IDRIS (2024-AD011013839R2 and AD011015497R1).

\bibliography{egbib}

\appendix
\setcounter{table}{6}
\setcounter{figure}{3}

\newpage
\section*{Appendix}

In this supplementary material, we describe the training protocols in~\cref{ssec:impl_details}, provide additional information about the used datasets in~\cref{ssec:datasets} and show per class results in~\cref{ssec:per_class_results}.

\section{Experimental setups, and implementation details}
\label{ssec:impl_details}
For further research, the code will be made publicly available at \begingroup \hypersetup{urlcolor=red}{\href{https://github.com/valeoai/muddos}{github.com/valeoai/muddos}}\endgroup.  Here we list out the training details. 

\subsubsection*{Training protocol for supervised training (3.1).}

For the supervised training we use standard augmentations on the point cloud: random rotation around the z-axis, random flip of the x or y axes, and scaling of the coordinates by a factor chosen uniformly at random within $[0.9, 1.1]$. 
The loss is the sum of the cross-entropy and the Lov\'asz loss \cite{lovasz}. We use AdamW as optimizer.
The backbones are trained for 45 epochs, using a batch size of $8$, a weight decay of $0.003$, a base learning rate of $10^{-3}$ with linear warmup during $4$ epochs and a cosine decay towards $0$.

\subsubsection*{Training protocol for multimodal distillation (3.2).} 
The lidar backbones are pretrained using ScaLR~\cite{puy2024three}. 
This pretraining method requires calibrated and synchronized cameras and lidars to establish correspondences between points and pixels.
It does not require manual annotations. The backbone is pretrained by minimizing the $\ell_2$ distance between normalized pixel-features (coming from the VFM) and normalized point-features (coming from the lidar backbone).
The pretraining hyperparameters are adapted from ScaLR. 
The weights of the lidar backbone are optimized using AdamW, a batch size of $2$, a weight decay of $0.03$, and a base learning rate of $5  \cdot  10^{\text{-}4}$ with a linear warmup phase followed by a cosine decay. The images are resized to $224 \times 448$. No augmentations are used on the images. We use the the point cloud augmentations of Sec.\textcolor{red}{3.1}.

\subsubsection*{Training protocol for downstream training (3.3).}
For (a), (b) and (c), the weights are optimized for 10 epochs of the source dataset, using a batch size of $8$, a base learning rate of $10^{\text{-}3}$ with linear warmup during $1$ epoch and a cosine decay reaching $0$ at the end of the last training iteration and a weight decay of 0.03.
For full finetuning (c), we also use a layerwise learning rate decay of $0.99$ \cite{balestriero2023cookbook,clark2020electra} and do not apply weight decay on the pretrained weights, as done in~\cite{puy2024three}.  

\subsubsection*{Training protocol self-training (3.5).}
For the self-training we continue to train only the MLP head and keep the weights of the backbone frozen. The teacher weights are a moving average of the student weights with momentum $0.99$. We alternate between a batch made of source point clouds and a batch made of target point clouds. 
The teacher provides pseudo class labels to the student for target point cloud. We keep the pseudo-label only when the corresponding softmax value is above $0.9$. 
The loss is the sum of the cross-entropy and Lov\'asz loss on source point clouds, and the cross-entropy on target point clouds.

\section{Datasets.}
\label{ssec:datasets}
We conduct our study using four datasets in total:
nuScenes~\cite{lidarseg_nuscenes}, SemanticKITTI~\cite{geiger2012cvpr, behley2019iccv}, Waymo Open Dataset~\cite{Ettinger_2021_ICCV}.
The pairs of source and target datasets considered are: (nuScenes, SemanticKITTI) denoted by N$\shortto$K, (SemanticKITTI, nuScenes) denoted by K$\shortto$N, (nuScenes, Waymo) denoted by N$\shortto$W, (Waymo, nuScenes) denoted by W$\shortto$N. It should be noted that for each dataset a different lidar sensor is used. In nuScenes the lidar used has $32$ beams, whereas both lidars in SemanticKITTI and Waymo have $64$ beams. This difference makes the settings especially difficult. The details of the used datasets are outlined in~\cref{tab:app:datasets}.
The performance is evaluated by computing the mIoU after re-mapping the original classes in each dataset to 10 common classes: `car', `bicycle', `motorcycle', `truck', `bus'/`oth. vehicle', `person', `driveable surface',  `sidewalk', `vegetation', and `terrain'. 

The exact class mapping, as also used in \cite{peng2025learning}, can be seen in~\cref{tab:app:class_mapping_ns_sk} and~\cref{tab:app:class_mapping_ns_wo}.

\section{Per class results.}
\label{ssec:per_class_results}

The per class results corresponding to the class mappings of~\cref{tab:app:class_mapping_ns_sk} and~\cref{tab:app:class_mapping_ns_wo} and the results in~\cref{tab:experiments_main_comparison} are presented in~\cref{tab:app:experiments_per_class_full}.

In~\cref{tab:app:experiments_per_class_ns_sk_full}, we also report the results of \method{} with the class mapping from~\cite{michele2024saluda}, this allows to compare to AdaBN~\cite{LI2018109}, PTBN~\cite{nado2020evaluating}, CoSMix~\cite{saltori2022cosmix} and SALUDA~\cite{michele2024saluda}.
Consistently with the results of the main paper, our method (the only one to be multimodal) surpasses the other approaches by a large margin ($+5.9\%$ mIoU) on N$\shortto$K.

\clearpage

\begin{table}[t]

\centering
\tabcolsep 1mm
\scalebox{0.85}{%
\begin{tabular}{l@{}lc|l|c|c|c|c|l}
\toprule
\rowcolor{violet!10}  
Dataset  & & & Lidar & \shortstack{Beams/\\Channels} & Cameras &cls. & \shortstack{Nb. Train/\\Val. Frames} & Region of the world
\\
\midrule
nuScenes & \cite{caesar2020nuscenes} &  (N) & HDL-32E & 32 & 6 & 16 &  28,130 / 6,019 & Boston, Singapore 
\\

SemanticKITTI & \cite{behley2019iccv} & (K) & HDL-64E & 64 & 1 & 19   & 19,130 / 4,071 & Karlsruhe 
\\
Waymo Open & \cite{Ettinger_2021_ICCV} & (W) & L.B.H. & 64 & 5  & 23 & 23,691 / 5976  & 3 US cities  \\
PandaSet$^\dagger$ & \cite{xiao2021pandaset} & (P) & Pandar64/ -GT & 64/150 & 6 & 37  & 3,800    & 2 US cities 
\\
\bottomrule
\\
\end{tabular}
}
\caption{\textbf{Datasets used in our domain adaptation experiments.} $^\dagger$: Only used for distillation.}
\label{tab:app:datasets}
\end{table}

\begin{table}[t]

\parbox{.48\linewidth}{

    \small
    \setlength{\tabcolsep}{2pt}
    \centering

        \begin{tabular}{c|c|c}
            \toprule
              
\rowcolor{violet!10}           
            \shortstack{Original\\nuScenes\\classes} & \shortstack{Shared\\set of\\classes} & \shortstack{Original\\SemanticKITTI\\classes} \\ 
             \bottomrule \vphantom{$X^{X^X}$}
            Car & Car &  Car\\
            \midrule
            Bicycle    & Bicycle & Bicycle$^\dagger$ \\
            \midrule
            Motorcycle & Motorcycle & Motorcycle$^\dagger$ \\
            \midrule
            Truck   & Truck & Truck \\
            \midrule
            Bus  &  Oth. vehicle & Oth. vehicle \\

            \midrule
            Pedestrian  & Pedestrian & Person \\
            \midrule
            \multirow{ 2}{*}{\shortstack{Driveable \\ Surface}}    & \multirow{ 2}{*}{\shortstack{Driveable \\ surface}} & Road,\\
            & & Parking \\
            \midrule
            Sidewalk   & Sidewalk &  Sidewalk \\
            \midrule
            Terrain    & Terrain &  Terrain\\
            \midrule
            Vegetation  & Vegetation & Vegetation, Trunk\\

            \bottomrule
            \multicolumn{1}{c}{}\\
        \end{tabular}
        \caption{
        \textbf{Class mapping used for N$\shortto$K and K$\shortto$N used in ~\cite{peng2025learning} and our work.} $^\dagger$: the classes `bicycle' and `motorcycle' do not include the original classes `motorcyclist' and `bicyclist', respectively, which are both mapped to `ignore', like all official classes not mentioned in this table.} 
        
        \label{tab:app:class_mapping_ns_sk}
}
\hfill
\parbox{.48\linewidth}{
    \small
    \setlength{\tabcolsep}{2pt}
    \centering

        \begin{tabular}{c|c|c}
            \toprule
              
\rowcolor{violet!10}           
            \shortstack{Original\\nuScenes\\classes} & \shortstack{Shared\\set of\\classes} & \shortstack{Original\\Waymo\\classes} \\ 
             \bottomrule \vphantom{$X^{X^X}$}
            Car & Car &  Car\\
            \midrule
            Bicycle    & Bicycle & Bicycle/ist \\
            \midrule
            Motorcycle & Motorcycle & Motorcycle/ist \\
            \midrule
            Truck   & Truck & Truck \\
            \midrule
            Bus  &  Bus &  Bus \\
            \midrule
            Pedestrian  & Pedestrian & Person \\
            \midrule
            \multirow{ 2}{*}{\shortstack{Driveable \\ Surface}}    & \multirow{ 2}{*}{\shortstack{Driveable \\ surface}} & Road,\\
            & & Lane marking \\
            \midrule
            Sidewalk   & Sidewalk &  Sidewalk \\
            \midrule
            Terrain    & Terrain &  Walkable\\
            \midrule
            Vegetation  & Vegetation & Vegetation, Tree Trunk\\

            \bottomrule
            \multicolumn{1}{c}{}\\
        \end{tabular}
        \caption{\textbf{Class mapping used for N$\shortto$W and W$\shortto$N in~\cite{peng2025learning} and our work.} All official classes in the original datasets not mentioned in this table are mapped to `ignore'.}
        \label{tab:app:class_mapping_ns_wo}
        
    ~\\~\\~\\
}
\end{table}

\begin{table}[t]
\small
\centering
\newcommand*\rotext{\multicolumn{1}{R{60}{1em}}}
\setlength{\tabcolsep}{3.5pt}
\begin{tabular}{ll|c|cccccccccc}
\toprule
{Method} 
    & Backbone
    & \rotext{mIoU\%} 
    & \rotext{Car} 
    & \rotext{Bicycle}
    & \rotext{Motorcycle} 
    & \rotext{Truck} 
    & \rotext{Bus/Oth. veh.}
    & \rotext{Pedestrian}
    & \rotext{Drive.~surf.}
    & \rotext{Sidewalk} 
    & \rotext{Terrain} 
    & \rotext{Vegetation}
\\

\midrule
\multicolumn{13}{l}{\textbf{N$\shortto$K}}
\\
Source only
& WI-768  
    & 44.6 & 89.9 & 4.1 & 	25.5 &	18.3 &	1.7 &	53.8 &	75.8 &	47.2 &	45 & 84.7\\
\rowcolor{orange!15}
Ours
& WI-768
    & 52.1 &89.7& 	19.7 &	48.0& 	26 & 	4.9 & 	63.2 &	76.2 &	43.6 &	61.3 & 	88.4   
\\
\midrule
\multicolumn{13}{l}{\textbf{K$\shortto$N}}
\\
Source only
& WI-768  
    & 55.1  & 75.9 & \textcolor{white}{0}5.1 & 48.4  & 38.2 & 38.9 & 55.2 & 90.0 & 53.2 & 56.5 & 89.8 
\\
\rowcolor{orange!15}
Ours
& WI-768
    & 66.4 & 83.2 & 10.5 & 74.4 & 57.0 & 58.8 & 67.9 & 91.6 & 60.0 & 70.1 & 91.0
\\
\midrule
\multicolumn{13}{l}{\textbf{N$\shortto$W}}
\\
Source only
    & WI-768
    &  37.1 & 49.5 & \textcolor{white}{0}2.1 & 17.3 & 21.5 & \textcolor{white}{0}9.2 & 41.3 & 72.3 & 43.1 & 39.7 & 75.4

\\
\rowcolor{orange!15}
Ours
    & WI-768
    & 69.1 & 85.1 & 49.5 & 48.9 & 41.1 & 75.9 & 81.2 & 88.9 & 61.8 & 66.6& 92.2
\\
\midrule
\multicolumn{13}{l}{\textbf{W$\shortto$N}}
\\
Source only
    & WI-768
     & 64.6 & 81.7 & 14.2 & 38.9 & 65.8 & 74.9 &  68.1 & 91.8 & 59.8 & 61.2 & 89.5

\\
\rowcolor{orange!15}
Ours
    & WI-768
    & 70.5& 80.8 & 29.7 & 75.8 &  71.1 &  71.7 & 68.9 & 91.1 & 55.0 & 69.2 & 91.8 
\\
\bottomrule
\\
\end{tabular}
\caption{
\textbf{Classwise IoU\% for N$\shortto$K, K$\shortto$N, N$\shortto$W and W$\shortto$N.}
The class mapping is the one used in \cref{tab:experiments_main_comparison} of the main paper.
}
\label{tab:app:experiments_per_class_full}
\end{table}

\begin{table}[t]
\small
\centering
\newcommand*\rotext{\multicolumn{1}{R{60}{1em}}}
\setlength{\tabcolsep}{2pt}
\begin{tabular}{llcc|c|cccccccccc}
\toprule
{Method}
    & \shortstack{Back-\\bone}
    & \shortstack{UDA\\Modal.}
    & {Norm.} 
    & \rotext{mIoU\%} 
    & \rotext{Car} 
    & \rotext{Bicycle}
    & \rotext{Motorcycle} 
    & \rotext{Truck} 
    & \rotext{Oth.~vehicle}
    & \rotext{Pedestrian}
    & \rotext{Drive.~surf.}
    & \rotext{Sidewalk} 
    & \rotext{Terrain} 
    & \rotext{Vegetation}
\\
\midrule
Source only$^\dagger$ 
    & MUNet.
    & - 
    & Batch 
    & 35.9 & 73.7 & 8.0 & 17.8 & 12.0 & 7.4 & 49.4 & 50.2 & 27.0 & 31.6 & 82.1
\\
Source only
    & WI-768
    & - 
    & Layer 
    & 44.6  & 89.8 & 4.2 & 25.6 & 18.3 & 1.6 & 53.8 & 75.8 & 47.2 & 45.0 & 76.8
\\
\midrule
AdaBN$^\dagger$ \cite{LI2018109}
    & MUNet.
    & U 
    & Batch 
    & 40.1  &  84.1 &  16.5 &  24.0 & 7.6 & 3.5 & 19.2 &  76.0 & 35.6  & 51.0 & 83.1
\\
PTBN$^\ddagger$ \cite{nado2020evaluating}
    & MUNet.
    & U 
    & Batch 
    & 39.4 & 	80.0 &	14.7 &	27.0 &	7.3 &	5.5 &	23.2	 &71.3 &	35.4 &	48.8& 	80.6
\\
CoSMix$^\dagger$ \cite{saltori2022cosmix}
    & MUNet.
    & U 
    & Batch 
    & 38.3
&77.1	&	10.4& 20.0	&{15.2}	&	6.6	&	{51.0}&	52.1&	31.8&	34.5&84.8
\\
SALUDA \cite{michele2024saluda}
    & MUNet.
    & U 
    & Batch 
    &  {46.2}
&{89.8} &	13.2&	26.2&	{15.3}	&	7.0	&	37.6&	{79.0}&	{50.4}&	{55.0}&{88.3} 
\\
\midrule
\rowcolor{orange!15}
Ours
    & WI-768
    & M 
    & Layer 
    & 52.1 & 88.9 & 19.8 & 48.1 & 26.0 & 5.0 & 63.2 & 76.2 & 43.6 & 61.3 & 88.5
\\
\bottomrule
\\
\end{tabular}
\caption{
\textbf{Classwise IoU\% for N$\shortto$K.} The results are obtained from \cite{michele2024saluda} for all methods marked with $^\dagger$, and from \cite{michele2024ttyd} for those marked with $^\ddagger$. Note that the class mapping in this table is the one used in \cite{michele2024saluda}, which has minor differences with the one used in \cref{tab:experiments_main_comparison}.
}
\label{tab:app:experiments_per_class_ns_sk_full}
\end{table}

\end{document}